\documentclass{article}
\usepackage{etoolbox}

\input{preamble-colm}
\usepackage{times}
\RequirePackage{caption}

\usepackage[utf8]{inputenc} %
\usepackage[T1]{fontenc}    %
\usepackage{hyperref}       %
\usepackage{url}            %
\usepackage{booktabs}       %
\usepackage{amsfonts}       %
\usepackage{nicefrac}       %
\usepackage{microtype}      %
\usepackage[dvipsnames]{xcolor}         %
\usepackage{wasysym}
\usepackage{siunitx}
\usepackage{lineno}

\definecolor{darkblue}{rgb}{0, 0, 0.5}
\hypersetup{colorlinks=true, citecolor=darkblue, linkcolor=darkblue, urlcolor=darkblue}

\title{Scaling Diverse Language Generation for 3D Visual Grounding}

\author{Austin T. Wang, Dongchen Yang \& Angel X. Chang \\
Simon Fraser University \\
\texttt{\{atw7,dongchen\_yang,angelx\}@sfu.ca} \\
{\small{\url{https://3dlg-hcvc.github.io/vigil3dpp/}}}
}

\begin{document}

\ifcolmsubmission
\linenumbers
\fi

\maketitle

\begin{figure}[ht]
  \vspace{-2mm}
  \centering
  \setkeys{Gin}{width=\linewidth}
  \begin{tabularx}{\textwidth}{>{\centering\arraybackslash}m{0.54\textwidth} @{\hspace{1mm}} Y}
    \includegraphics{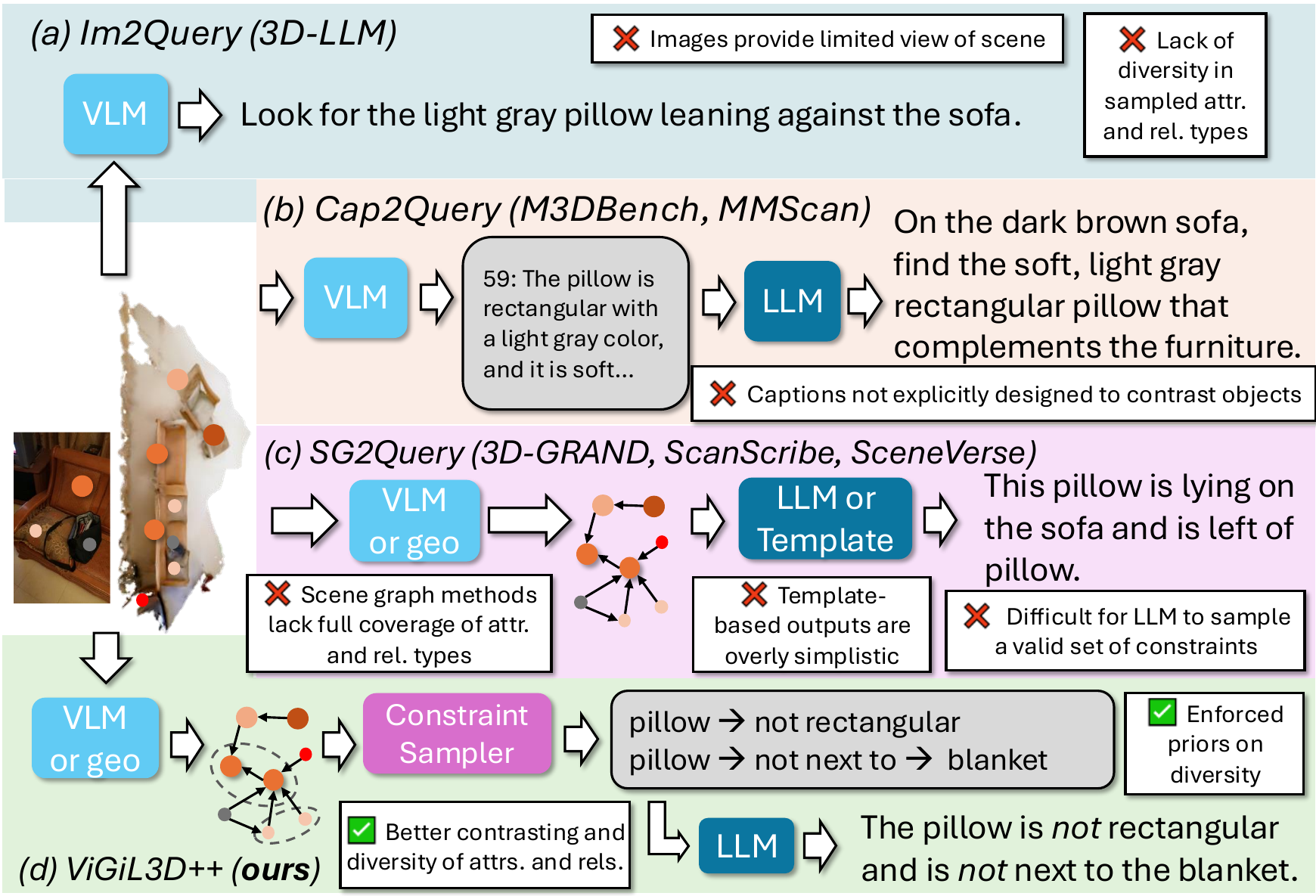} & \includegraphics{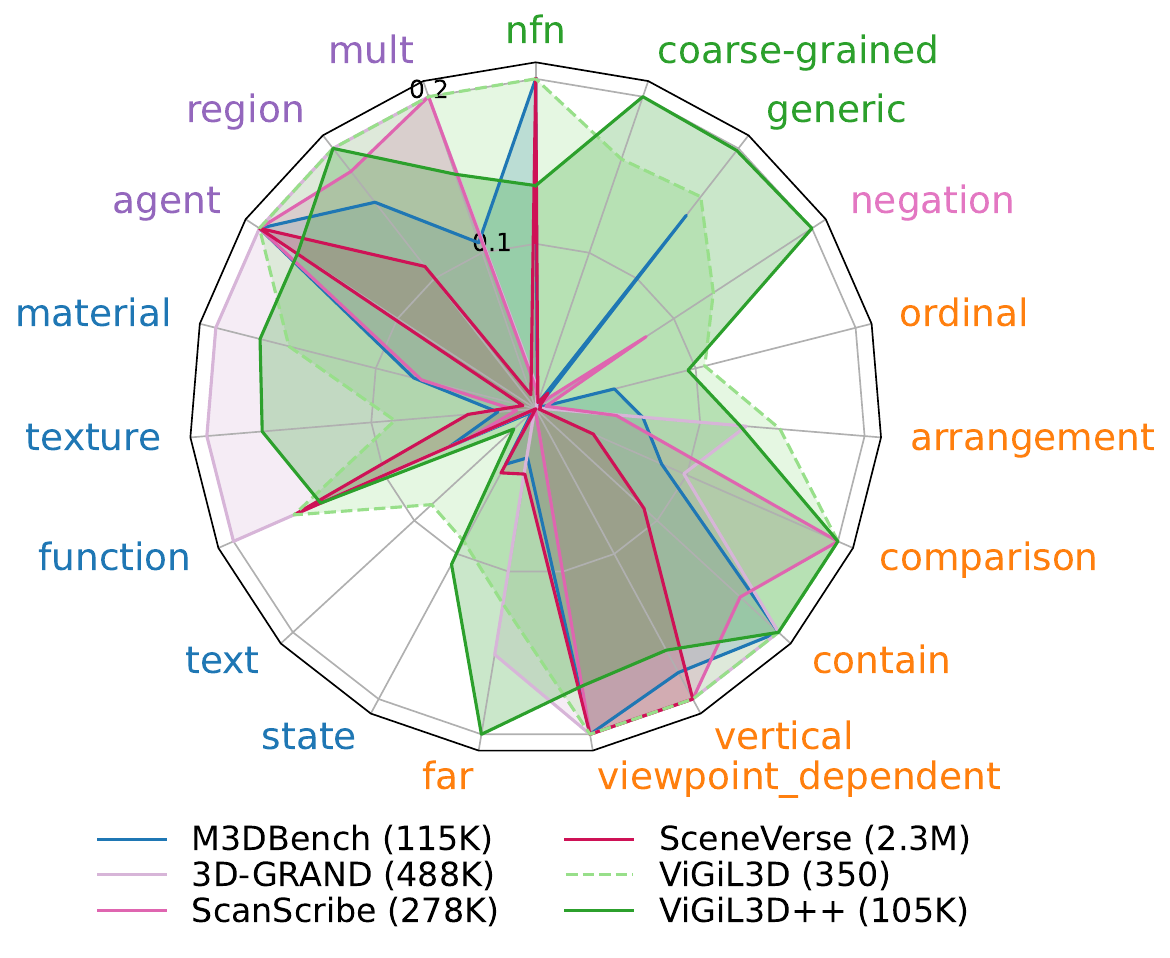}
  \end{tabularx}
\captionof{figure}{
  \textbf{Comparison of 3D query generation methods.} 
  \textit{Left}: 
  To scale datasets for visual grounding (VG), recent work automatically generate VG descriptions (queries) for objects in 3D scene datasets using templates or direct MLLM-based generation from \textcolor{teal}{(a) images}, \textcolor{orange}{(b) captions}, or \textcolor{CarnationPink}{(c) scene graphs}. 
  Their queries lack linguistic diversity and fail to synthesize information across the \emph{entire} scene into cohesive sentences. We propose \textcolor{forestgreen}{(d) \oursshort}, which applies \emph{cross-referencing} to increase attribute consistency among similar objects and overall validity in queries and \emph{constraint sampling} to enforce greater diversity in the targets and constraints.
  \textit{Right}: \textcolor{forestgreen}{\oursshort} has better representation of linguistic patterns than existing scaled VG datasets, which are strong in only a few categories. \oursshort can also scale without manual annotation, unlike ViGiL3D~\citep{wang2024vigil3d}.
}
  \label{fig:teaser}
\end{figure}

\begin{abstract}
    Developing robust models for 3D visual grounding (3DVG), the localization of entities in a 3D scene described in natural language, is important for enabling agents to correspond spatial language with objects in the physical world.
    However, the lack of diverse descriptions at scale prevents models from generalizing beyond simple linguistic patterns.
    Recent such attempts lack diversity in the constraint types and language used to ground objects. Captioning methods cannot precisely contrast objects, which is important for visual grounding.
    We therefore propose \oursshort, a scalable, scene-agnostic method that generates diverse visual grounding queries by combining constraint sampling in scene graphs with the language generation of LLMs. We show that it has greater diversity over existing scaled datasets and improves model performance over several 3DVG benchmarks but also illuminates outstanding limitations of VLMs.

\end{abstract}

\section{Introduction}
\label{sec:intro}

  Embodied agents that understand, navigate, and manipulate 3D environments have broad applications in robotics, assistive technology, and augmented reality. 
  One key to enabling these capabilities is 3D visual grounding (3DVG), which involves localizing an object described by natural language in a scene. 
  To achieve this, agents must handle the varied ways objects and their relationships can be described. 
  However, existing large-scale 3DVG datasets lack linguistic diversity, and thus models trained on these datasets do not generalize well beyond simple descriptions. Building a diverse 3DVG dataset is key for agents to understand the full referring expression space for 3D scenes.

  The established datasets for 3DVG~\citep{chen2020scanrefer, achlioptas2020referit3d} capture common human descriptions but are not suitable for large-scale training. They are also limited in the set of visual cues and language used to disambiguate objects. 
  ViGiL3D~\citep{wang2024vigil3d} is diverse but lacks scale due to the manual annotation required.
  Recent work has developed LLM-based pipelines for scaling grounding annotations~\citep{yang20243d, zhu20233d, jia2024sceneverse}. 
  However, such pipelines only describe one object in the scene and use simple template-based constructions or verbose captions to minimize errors or ambiguities. 
  Neither strategy generates descriptions that fully capture the diversity of language queries (\cref{fig:teaser} right). 
  Thus, using VLMs to create a diverse dataset similar to ViGiL3D while achieving scalability and high fidelity in generated queries is still an underexplored challenge. In this paper, we investigate \textbf{\emph{how capable VLMs are of generating valid, diverse visual grounding descriptions}}.

  To answer this question,
  we reimplement common paradigms for generating object descriptions from images, captions, and scene graphs directly using VLMs (\cref{fig:teaser}). These methods generate queries with low validity and diversity because, like existing methods, they often output verbose descriptions and either have limited scope of the scene (image- and caption-based) or struggle to synthesize relationships in the scene graph into cohesive sentences (scene graph-based).
  We improve on these methods by proposing \textbf{\oursshort}, a scalable pipeline that samples constraints from the scene graph and uses LLMs to rephrase them into linguistically diverse and high-fidelity 3DVG queries (\eg in a scene with four chairs, describing \textit{the two accent chairs not to the right of the table}). We innovate in two key ways.
  First, our solution uses \emph{cross-referencing} to enforce consistency in the terms used to describe similar objects (\eg. if one of two similar chairs is described as an \textit{accent chair}, then the other is also). This ensures that objects are contrasted due to differences in properties rather than terminology.
  Secondly, we use constraint sampling to enforce diversity in the constraint types and to construct the logical constraints to differentiate objects, thus narrowing the responsibility of the LLM to language generation.
  \oursshort generates queries with higher linguistic diversity and validity rates than previous methods. With data generated by \oursshort, we show that we can train models that perform better on linguistically diverse 3DVG benchmarks. 
  However, we also find that there are outstanding limitations of VLMs in generating scene graphs and phrasing 3DVG descriptions, highlighting areas for future improvement for spatial reasoning and language generation.

  In summary, we make the following contributions:
  \begin{itemize}[leftmargin=*]
    \item We analyze the validity and linguistic diversity of VLM-based approaches, showing the limitations of VLMs and benefits of our method for the quality and diversity of visual grounding descriptions.
    \item We propose \oursshort, a pipeline for generating more diverse and higher-fidelity visual grounding descriptions compared to prior scaled 3DVG datasets.
    \item We propose a scene graph extraction method for 3DVG query generation by extracting attributes and relationship types with \emph{cross-referencing} to enforce consistency of descriptions across objects.%
    \item We show that training on \oursshort enables a competitive baseline for dense multi-target 3D visual grounding, outperforming comparable architectures across multiple 3DVG benchmarks.
  \end{itemize}

\begin{table*}
\centering
\resizebox{\linewidth}{!}
{
\begin{tabular}{@{}l r c c c c c c c c r c @{}}
\toprule
Dataset                                         & VG    & method     & sg      & zero         & mult         & dense        & diverse      & auto         & analysis     & val   & avail \\
\midrule
MMScan~\citeyearpar{lyu2024mmscan}              & 1.28M & sg         & -       & \myxmark     & $\sim$       & \mycheckmark & \myxmark     & \myxmark     & $\sim$       & *95.0  & \myxmark     \\
ScanScribe~\citeyearpar{zhu20233d}              & 278K  & sg         & GT      & \myxmark     & \myxmark     & \myxmark     & \myxmark     & \myxmark     & \myxmark     & -      & \mycheckmark \\
LEO-Align~\citeyearpar{huang2023embodied}       & 354K  & sg         & 3DSSG   & \myxmark     & \myxmark     & \myxmark     & \myxmark     & \myxmark     & $\sim$       & *100.0 & \mycheckmark \\
SceneVerse~\citeyearpar{jia2024sceneverse}      & 2.3M  & sg         & VLM,geo & \myxmark     & \myxmark     & \myxmark     & \myxmark     & \myxmark     & \myxmark     & *96.9  & \mycheckmark \\
\midrule
ScanReason~\citeyearpar{zhu2024empowering}      & 13K   & cap        & -       & \myxmark     & \myxmark     & \myxmark     & \myxmark     & \mycheckmark & \myxmark     & -      & \myxmark     \\
3D-LLM~\citeyearpar{hong20233d}                 & 75K   & im         & -       & \myxmark     & \myxmark     & \myxmark     & \myxmark     & \mycheckmark & \myxmark     & *63.3  & $\sim$ \\
M3DBench~\citeyearpar{li2023m3dbench}           & 115K  & cap        & -       & \myxmark     & \mycheckmark & \myxmark     & \myxmark     & \mycheckmark & $\sim$       & -      & \mycheckmark \\
EmbodiedScan~\citeyearpar{wang2024embodiedscan} & 970K  & sg         & geo     & \myxmark     & \myxmark     & \myxmark     & \myxmark     & \mycheckmark & $\sim$       & -      & \mycheckmark \\
3D-GRAND~\citeyearpar{yang20243d}               & 488K  & sg         & VLM     & \myxmark     & \myxmark     & \mycheckmark & \myxmark     & \mycheckmark & $\sim$       & 38.3   & \mycheckmark \\
\rowcolor{gray!20}\oursshort                    & 205K  & sg+solver  & VLM,geo & \mycheckmark & \mycheckmark & \mycheckmark & \mycheckmark & \mycheckmark & \mycheckmark & 57.1   & \mycheckmark \\
\bottomrule
\end{tabular}
}
\caption{\textbf{Comparison of MLLM-generated 3D-VL datasets}.
We indicate the number of \textbf{VG} prompts for each dataset and the \textbf{method} (\textbf{im}2query, \textbf{cap}2query, or \textbf{sg}2query) to extract the prompts.
For those using \textbf{s}cene \textbf{g}raphs, we indicate the graph source (GT) or method used (3DSSG~\citep{wald2020learning}, VLM, or \textbf{geo}metric reasoning).
We also indicate which work create \textbf{zero-} or \textbf{mult}i-target queries, provide \textbf{dense} alignment, generate \textbf{diverse} queries (see \autoref{tab:dataset-comparison}), are fully \textbf{auto}mated, have linguistic \textbf{analysis}, and have \textbf{avail}able implementations.
\mycheckmark means ``yes'', \myxmark means ``no'', and $\sim$ means ``partial''.
\oursshort generates zero-target queries and is among the few with multi-target queries and dense annotations. \oursshort has diverse prompts and extensive comparison of linguistic diversity. Methods with higher \textbf{val}idity require manual validation.
*As reported by existing works.
}
\label{tab:pipeline-comparison}
\vspace{-6mm}
\end{table*}

\section{Related Work}

  \textbf{Indoor 3DVG Datasets.}
  3DVG indoor datasets are defined by the original scene dataset and the grounding descriptions annotated on those scenes. Scene datasets range from real-world reconstructions \citep{dai2017scannet, yeshwanth2023scannet++, Wald2019RIO, baruch2021arkitscenes, ramakrishnan2021hm3d, mao2022multiscan, chang2017matterport3d} to synthetic datasets \citep{fu20213d, zheng2020structured3d, deitke2022procthor, khanna2023hssd}, including single-room~\citep{dai2017scannet} and multi-room scenes~\citep{yeshwanth2023scannet++}. 
  ScanNet~\citep{dai2017scannet} is among the most popular scene datasets for 3D language annotations, though more recent datasets~\citep{yeshwanth2023scannet++, baruch2021arkitscenes} have higher quality reconstructions and RGB-D videos. 
  Many grounding datasets were created through crowdsourced annotation.
  ScanRefer, Multi3DRefer, Nr3D, and Sr3D~\citep{chen2020scanrefer, achlioptas2020referit3d, kato2023arkitscenerefer, zhang2023multi3drefer} are among the most common, but they are small, suffer from low-quality scenes, and lack diverse language patterns. 
  More recent datasets attempted to expand the diversity and quality of descriptions~\citep{he2024segpoint, delitzas2024scenefun3d, wang2024vigil3d, huang2025unveiling}, such as by emphasizing function, reasoning, or more complex language parsing, but they rely on manual annotations and are thus hard to scale.

  \textbf{VLM-scaled 3D Datasets.}
  Recent work has leveraged LLMs and MLLMs to generate 3DVG queries on aggregate scene datasets, typically covering multiple 3D language tasks including visual question-answering~\citep{hong20233d, zhang2024vla, yang2024synvl3d, zhang2025open3d} and visual grounding~\citep{jia2024sceneverse, yang20243d, zhu20233d, zhan2024mono3dvg, lyu2024mmscan, huang2023embodied, li2023m3dbench}.
  One strategy~\citep{zhu20233d, jia2024sceneverse, hong20233d} is to incorporate existing manual datasets and rephrase with LLMs to expand the linguistic variation, but these are limited to the targets and constraint types already represented.
  Others~\citep{xian2023towards, wang2023robogen, hong20233d, katara2024gen2sim, zhu2024empowering, chang2024partnr, zhao2025his, park2025nuplanqa} directly prompt the LLM with object information encoded as structured text, relying on the LLM to implicitly parse the relationships between objects.
  Most methods~\citep{yang20243d, jia2024sceneverse, zhu20233d, lyu2024mmscan, huang2023embodied, zhan2024mono3dvg, zhang2024task}, however, use scene graphs to build an explicit representation of the scene, including attributes of and relationships between objects.
  Such methods either limit the domain of descriptions or output verbose descriptions to eliminate ambiguities to produce valid descriptions, resulting in low diversity. Our method achieves higher validity and diversity by explicitly sampling constraints on scene graphs to enforce logical consistency, while still using LLMs to generate natural descriptions. We compare \ourdataset to existing methods in \cref{tab:pipeline-comparison}.

  \textbf{Open-Vocabulary 3DVG Models.}
  While traditional 3DVG models are trained on ScanRefer~\citep{chen2020scanrefer} and Nr3D~\citep{achlioptas2020referit3d}, scaled 3DVG datasets and general-purpose MLLMs have enabled training open-vocabulary 3DVG models. Transformer-based models~\citep{peng2023openscene, zhu20233d, zhu2024unifying, jia2024sceneverse, yu2025inst3d} first encode point cloud or RGB-D features and text features independently, using a strong pretrained language encoder and large pretraining datasets for open-vocabulary performance. They then fuse features together through an attention-based network to make a final prediction of the described target. Other recent methods use pretrained or finetuned LLMs to predict object targets based on structured text~\citep{yang20243d, yuan2024visual, li2025seeground}. We use a similar architecture to \cite{zhu20233d} to show the power of our dataset but leverage the dense annotations of \ourdataset for additional training objectives.

\section{\oursshort Pipeline}

    \begin{figure*}[t]
  \centering
  \includegraphics[trim={0 0 0 4px},clip,width=\linewidth]{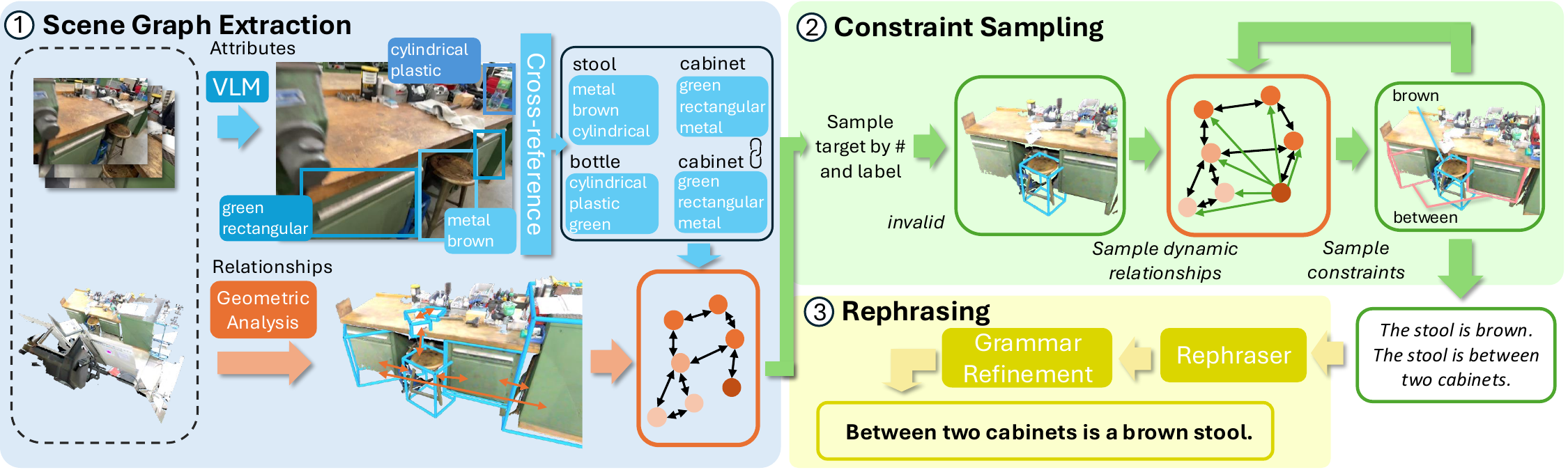}
  \caption{\textbf{Grounding annotation pipeline.} 
  We present our pipeline for sampling grounding descriptions from 3D scenes. 
  \textbf{1)} We extract a \sgprop scene graph by parsing attributes using a VLM from multi-view images of each object and relationships through geometric analysis.
  \textbf{2)} For each query, we sample targets conditioned on the query type (zero-, single-, or multi-target) and label. 
  Constraints are iteratively sampled from the scene graph until the targets are appropriately constrained. 
  \textbf{3)} We use an LLM to rephrase the templated constraints into a natural language query.
  }
  \label{fig:llm-based-pipeline}
 \vspace{-2mm}
\end{figure*}

    \begin{figure*}[t]
  \centering
  \includegraphics[trim={0 0 0 0},clip,width=\linewidth]{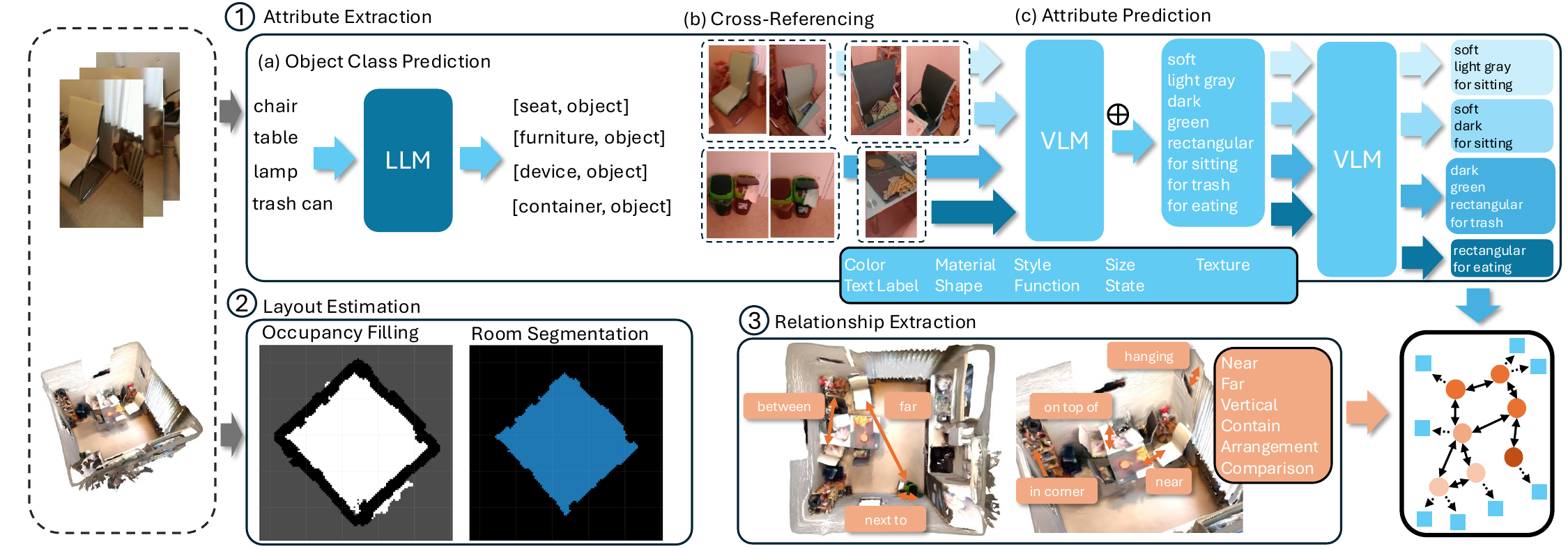}
  \caption{\textbf{Scene graph extraction pipeline.} We extract the scene graph from the point cloud and RGB images. We first initialize the objects from ground truth or predicted segmentation. We then add attributes and relationships to the graph. \textbf{1)} To extract attributes, (a) we first augment the initial object labels with additional object classes. (b) We then identify identical objects, and subsequently we perform (c) \emph{attribute prediction} and assignment using a VLM. \textbf{2)} \emph{Layout estimation} provides the room layout of the scene for determining the relationships of objects relative to the room, and \textbf{3)} \emph{relationship extraction} identifies spatial relationships based on geometry.
  }
  \label{fig:sg-pipeline}
\end{figure*}

  We design a fully automated pipeline that
  generates a \emph{diverse} set of 3DVG queries given a point cloud and RGB images of a scene (\cref{fig:llm-based-pipeline}). 
  Our method improves existing ones by 
  enforcing consistency in attribute assignment through \emph{cross-referencing}
  and leveraging \emph{constraint sampling} to enforce diverse sampling and ensure logical consistency. 
  
  Specifically, our grounding annotation pipeline has three stages: scene graph extraction, constraint sampling, and rephrasing constraints into natural language (\cref{fig:llm-based-pipeline}).
  We first define the set of potential constraints using semantic scene graphs (\cref{sec:sg-extraction}) representing objects in the scene, their attributes (\eg color, shape), and the spatial relationships between them (\eg near, above). 
  We then iteratively sample constraints, or attributes and relationships, from the scene graph, reducing the pool of potential targets until only the target objects remain (\cref{sec:constraint-algorithm}).
  After constructing a set of constraints, we use an LLM to rephrase them into natural language to obtain rich, diverse descriptions (\cref{sec:rephrasing}).
  Compared to prior approaches for generating 3DVG data, our approach ensures that the sampled constraints uniquely identify the target object(s) given the scene graph and controls the distribution of target object counts, constraint types (\eg attributes, relations, negation), and object referents. %

  \subsection{Scene Graph Extraction}
  \label{sec:sg-extraction}
  \vspace{-5pt}
  Given the high complexity of 3D scenes, we extract a scene graph to produce a semantic-focused representation by identifying the objects in the scene, similar objects and attributes for each object, and relationships between objects (\cref{fig:sg-pipeline}). 
  We improve existing 3D scene graph extraction methods by \emph{cross-referencing} across groups of similar objects to assign attributes consistently across objects. We also expand the attribute and relationship types extracted, including egocentric directional (\eg \textit{left} or \textit{right} relative to an agent in the scene) and ordinal relationships.
  We outline our method below and include details in \cref{appendix:sg-extraction}.

  \mypara{Objects.}
  As there is extensive work on 3D object detection and segmentation~\citep{guo2020deep,zhu2024survey}, we assume the segmentation is available and focus on expanding the object labels.
  We use ground-truth object segmentations when available, as in prior work~\citep{zemskova20253dgraphllm}, to determine the labels and masks of the point cloud for each object. 
  For scenes without ground-truth segmentations, objects can be predicted using segmentation methods such as Mask3D~\citep{schult2023mask3d}. 
  We prompt the LLM to predict additional object classes given the initial object classes to augment each object with labels of varying granularity. 
  
  \mypara{Attributes.} 
  We extract attributes by \emph{cross-referencing} between objects for consistency. 
  We prompt the VLM to identify \emph{similar} objects that have the same attributes (\ie same color, size, material, and other attributes) given views of two objects of the same class.
  For each group of similar objects, we query the VLM twice with views of one object in the group, first to retrieve a list of attributes and their respective types per object and second to augment from the full list of predicted attributes. This ensures identical attributes among similar objects and consistent terminology across different objects.
  
  \mypara{Relationships.} 
  We improve existing methods by expanding the types of supported relationships. In addition to \emph{query-independent} relationships, which are true regardless of context (\eg near, above), we generate \emph{query-dependent} relationships that are relative to an agent in the scene or a subset of the objects (\eg left of an agent or the tallest among furniture).
  We predict relationships using geometric and VLM-based methods, The latter are predicted during constraint sampling.
  To restrict relationships to within rooms and ground objects in their rooms (\eg in the corner or center), we predict room boundaries by identifying connected cells in the $xy$ grid that are separated by walls.

  \subsection{Constraint Sampling}
  \label{sec:constraint-algorithm}
  \vspace{-5pt}
  We implement a constraint sampler to select the \emph{constraints}---attributes or relationships---used in each description. Unlike simple template construction, constraint sampling allows us to enforce priors on the distribution of constraint types and preserve logical consistency between the scene graph and generated description, allowing for more complex outputs. 
  We first randomly select a subset of targets from the scene and then iteratively sample constraints from the graph to reduce the set of possible solutions to the selected targets. There are four key steps in our implementation: target selection, constraint space definition, sampling, and the terminal condition.

  \mypara{Target selection.}
  We condition the selection of target objects $T$ on the number of targets (zero-, single-, or multi-target) and object class to ensure there are sufficient numbers of each type of query. For zero-target queries, we generate a description of an object of a sampled \emph{common} class that does not describe any object in the scene.
  For single-target queries, we balance between unique and common classes to ensure that knowing the target class is insufficient in most queries to predict the target.
  Lastly, for multi-target queries, we initialize the target to all objects of a common class, which will be reduced during sampling.

  \mypara{Constraint pool construction.} To build the pool of constraints at each iteration, we first sample a \emph{referent} (the target or anchor object referenced in a previously sampled constraint). To minimize ambiguity, referents with many distractor objects are prioritized for exploration.
  We then enumerate all constraints on the referent that reduce the solution space toward the intended anchors or targets. For zero- and single-target queries, we aim to eliminate distractor objects, whereas for multi-target queries, we aim to reduce the number of targets (while keeping at least two targets). By dynamically choosing the set of targets for multi-target queries, we keep all targets semantically related.
  We further augment the static constraints from the scene graph with query-dependent relationships based on a sampled agent position and the given referent. Further details can be found in \cref{appendix:sg-method-rel-dependent}.

  \mypara{Sampling.}
  Given a pool of constraints, we balance language patterns in the dataset by weighing the sampling toward uniform frequencies of each constraint type, first choosing 1) attributes or relationships, 2) the type of constraint (\eg color or size for attribute), 3) the use of negation, and 4) the constraint itself. 
  To perform sampling, we prioritize constraints that reduce the distractor pool by at least half and, for relationships, weigh constraints whose referents have fewer distractors higher to keep the solution space from exploding. We also weigh probabilities against the sampling history, since not all constraint types appear with equal frequency.
  We repeat pool construction and sampling until the terminal condition is reached.

  \mypara{Terminal condition.}
  For zero- and single-target queries, we terminate sampling when the constraints resolve to an empty solution or the desired target, respectively. 
  For multi-target queries, to restrict the target space to meaningfully related objects, we sample at least one constraint with probability $p_{\text{init}}$ and subsequently terminate sampling probabilistically, with a higher weight as the number of targets decreases. The mathematical formulation can be found in \cref{appendix:terminal-condition}.

  Ultimately, constraint sampling yields a set of constraints that is consistent with the scene graph and traceable in its solution construction. Furthermore, our solution gives the user control over the frequencies of different features of the queries to enforce semantic diversity over the types of objects and constraints described.

  \subsection{Rephrasing}
  \label{sec:rephrasing}

  Given the sampled constraints, we use an LLM to transform templated versions of those constraints into a candidate grounding description.
  To solicit linguistic diversity from LLMs, we provide manually constructed in-context examples designed to represent a wide range of language patterns.
  Additionally, we prompt the LLM to preserve ID tags from the sampled constraints, allowing for dense annotations of the referred objects in the description.
  We then pass the generated queries back to the LLM to correct grammatical errors and ambiguous phrasing.
  This allows \oursshort to achieve natural phrasing while preserving the semantic content of the original constraints.

\section{\oursshort Dataset}
  \label{sec:test-dataset}

  \begin{table*}[tb]
\centering
\resizebox{\linewidth}{!}
{
\begin{tabular}{@{}llrrrrrrrr@{}}
\toprule
& scenes & \# scenes & \# desc. & vocab & sent. & length & zt & st & mt \\
\midrule
\ourdataset & ScanNet, 3RScan, MultiScan & 1140 & 105K & 8.6K & 1.3 & 19.6 & 15.1K & 70.8K & 18.7K \\
\ourdatasetlarge & ScanNet, 3RScan, MultiScan & 1140 & 206K & 17.0K & 1.4 & 17.9 & 20.9K & 154K & 31.2K \\
\bottomrule
\end{tabular}
}
\caption{\textbf{Statistics of \oursshort.} We generate \ourdataset on ScanNet~\citep{dai2017scannet}, 3RScan~\citep{Wald2019RIO}, and MultiScan~\citep{mao2022multiscan} scenes. We further combine \ourdataset with ScanRefer~\citep{chen2020scanrefer} and Multi3DRefer~\citep{zhang2023multi3drefer} (\ourdatasetlarge) to leverage existing data. We report the source and number of scenes and \textbf{desc}riptions and the \textbf{vocab}ulary size. We additionally include the average number of \textbf{sent}ences and words (length) per description. Lastly, we report the number of \textbf{z}ero-, \textbf{s}ingle-, and \textbf{m}ulti-\textbf{t}arget descriptions.
}
\label{tab:dataset-overview}
\end{table*}

  We showcase the capabilities of \oursshort for training by generating a large multi-scene dataset, demonstrating its scalability along both scenes and descriptions. We additionally combine \ourdataset with ScanRefer~\citep{chen2020scanrefer} and Multi3DRefer~\citep{zhang2023multi3drefer} (\ourdatasetlarge) to evaluate the complementary impact of our data with manual annotations. We show the statistics of the main datasets created in \autoref{tab:dataset-overview} and compare our dataset to existing datasets in \autoref{tab:all-dataset-overview-general} and \autoref{tab:all-dataset-overview-example}. 
  
  We use GPT-4.1~\citep{achiam2023gpt} as our VLM, running on one L40S or A40 GPU with 32 GB of CPU RAM. Generating \ourdataset required around 6 machine-days of runtime, including roughly 2-5 minutes per scene graph and 3.7 seconds per 3DVG description.
  See \cref{appendix:dataset} for pipeline parameter settings and ablations on the choice of VLM. 

\section{\oursmodel}
\label{sec:model}

  \begin{figure*}[t]
  \centering
  \includegraphics[trim={0 0 0 4px},clip,width=\linewidth]{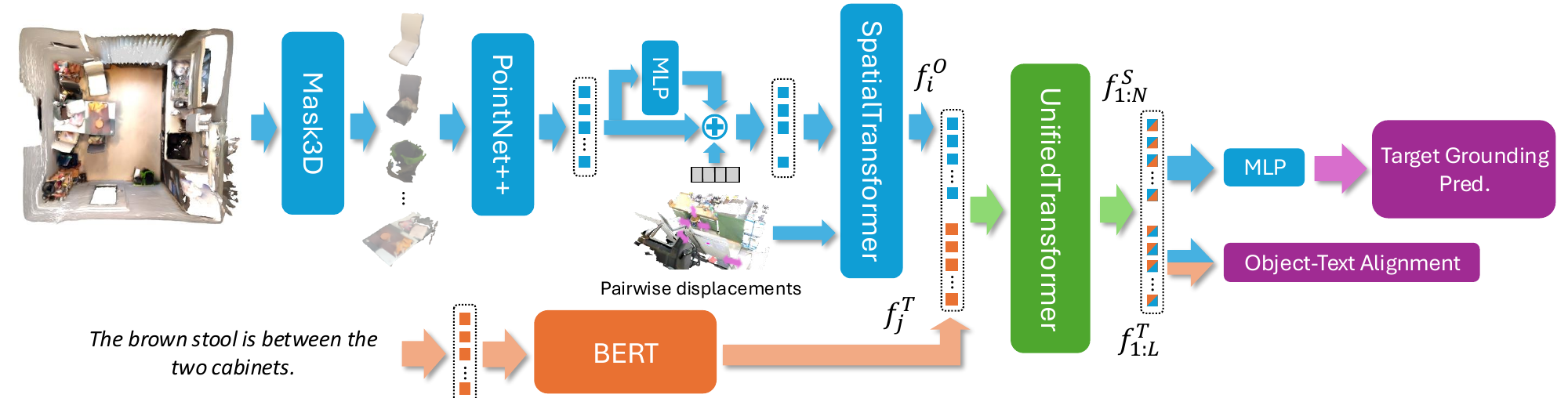}
  \caption{\textbf{Model architecture.} Given a grounding description and a point cloud, the point cloud is first segmented, using GT during training and either GT or Mask3D~\citep{schult2023mask3d} predictions during inference, and then processed using PointNet++~\citep{qi_pointnet_2017} to extract point features per object. We then add explicit pairwise spatial information to output object features $f_i^O$. For text, we use a BERT-like~\citep{devlin2019bert} module to extract text features $f_j^T$. We pass the tokens to a unified transformer to align across modalities, predicting the final target off the fused object features.
  }
  \label{fig:model-architecture}
 \vspace{-3mm}
\end{figure*}

To study the benefit of training on \oursshort, we implement a baseline architecture \oursmodellong following prior work~\citep{chen2022language, zhu20233d}.
We take as input a point cloud, segmented by ground truth or 3D object segmenters such as Mask3D~\citep{schult2023mask3d} into objects $\{O_i\}$.
Given the segmented point cloud and a grounding description, we obtain features $\{f_i^O\}$ for each object and language features $\{f_j^T\}$ for text tokens.  
The features are combined using a transformer into fused object features $f_{1:M}^S$ and language features $f_{1:t}^T$.  See \cref{fig:model-architecture} for an overview of the architecture.
To exploit the key characteristics of \oursshort, we extend the base model with the following modifications. For details on training losses and ablations, see \cref{appendix:model}.

\mypara{\textit{Generalized object features.}} While Vil3DRel and 3D-VisTA assume that the object classes are in ScanNet200 and use GloVe~\citep{pennington2014glove} word vectors to construct the object features, we replace them with learned semantic embeddings to generalize to any object classes.

\mypara{\textit{Dense alignment.}} We leverage the dense annotations of all referenced objects in \oursshort to compute a binary cross-entropy loss against a cross-attention map between text tokens and objects. Additionally, we apply an MLP to the object features to predict which objects in the scene are anchors.

\mypara{\textit{Variable-target predictions.}} While 3D-VisTA was designed to only predict a single target object, we follow M3D-CLIP~\citep{zhang2023multi3drefer} and apply a binary cross-entropy loss across all objects to predict a variable number of targets.

\section{Experiments}
  \label{sec:experiments}
  
  We investigate three aspects of the descriptions generated by \oursshort: validity (how often the queries correctly describe their targets), linguistic diversity, and value as training data for 3DVG.

\begin{table}[tb]
\centering
{
\begin{tabular}{@{}crrrrrr@{}}
\toprule
& Zero & \multicolumn{3}{c}{Single} & Multi & Total \\
\cmidrule(){2-2} \cmidrule(l){3-5} \cmidrule(l){6-6} \cmidrule(l){7-7} 
Dataset & & Unique & Common & Total & & \\
\midrule
Im2Query & - & 68.0 (18.3) & \best{44.0 (19.5)} & \best{56.0 (13.8)} & - & 56.0 (13.8) \\
Cap2Query & - & 87.0 (13.8) & 25.9 (16.5) & 54.0 (13.8) & - & 54.0 (13.8) \\
SG2Query & \best{80.0 (9.7)} & 91.7 (15.6) & 29.2 (12.9) & 41.7 (12.5) & 8.0 (10.6) & 52.7 (8.0) \\
3D-GRAND & - & 51.9 (8.6) & 12.9 (7.8) & 38.3 (6.7) & - & 38.3 (6.7) \\
\rowcolor{gray!20} \oursshort & 78.6 (9.6) & \best{100.0 (0.0)} & 43.1 (12.0) & 47.1 (11.7) & \best{45.7 (11.7)} & \best{57.1 (6.7)} \\
\bottomrule
\end{tabular}
}
\caption{\textbf{Query validation.} We run a user study to evaluate the validity of generated queries for \oursshort and baselines. For each type, we measure the proportion of queries that describe only the associated targets (or no targets for zero-target queries). For single-target queries, we split validity by targets that are \emph{unique} in their object class vs. \emph{common}. The 95\% CI is shown in parantheses.}
\label{tab:prompt-validation}
\end{table}

\begin{table*}[tb]
\centering
\resizebox{\linewidth}{!}
{
\begin{tabular}{@{}l rrrcc rrrcccc ccc ccc cr @{}}
\toprule
& \multicolumn{5}{c}{Attributes} & \multicolumn{7}{c}{Relationships} & \multicolumn{3}{c}{Target Reference} & \multicolumn{3}{c}{Anchor Type} & \multicolumn{2}{c}{Language} \\
\cmidrule(){2-6} \cmidrule(l){7-13} \cmidrule(l){14-16} \cmidrule(l){17-19} \cmidrule(l){20-21}
Dataset & all & tgt & anc & func & state & all & tgt & anc & far & arr & ord & comp & gen & CG & NFN & mul & non & agt & neg & 2lex \\
\midrule
\oursshortman~\citeyearpar{wang2024vigil3d} & 1.62 & 1.09 & 0.53 & \mycheckmark & \mycheckmark & 1.82 & 1.46 & 0.35 & \mycheckmark & \mycheckmark & \mycheckmark & \doublecheckmark & \mycheckmark & \mycheckmark & \doublecheckmark & \doublecheckmark & \doublecheckmark & \doublecheckmark & \mycheckmark & \best{0.45} \\
\midrule
M3DBench~\citeyearpar{li2023m3dbench} & 1.86 & 1.48 & 0.39 & \mycheckmark & \myxmark & 1.37 & 1.23 & 0.14 & \myxmark & \mycheckmark & \myxmark & \mycheckmark & \mycheckmark & \myxmark & \doublecheckmark & \mycheckmark & \mycheckmark & \doublecheckmark & \myxmark & 0.09 \\
3D-GRAND~\citeyearpar{yang20243d} & 5.81 & 4.68 & 1.12 & \doublecheckmark & \myxmark & 2.81 & 2.71 & 0.10 & \mycheckmark & \mycheckmark & \myxmark & \mycheckmark & \myxmark & \myxmark & \myxmark & \doublecheckmark & \doublecheckmark & \doublecheckmark & \myxmark & 0.05 \\ %
ScanScribe~\citeyearpar{zhu20233d} & 6.04 & 1.15 & 4.89 & \myxmark & \myxmark & 3.55 & 3.35 & 0.21 & \myxmark & \myxmark & \myxmark & \doublecheckmark & \myxmark & \myxmark & \myxmark & \doublecheckmark & \mycheckmark & \doublecheckmark & \myxmark & 0.10 \\
SceneVerse~\citeyearpar{jia2024sceneverse} & 0.41 & 0.35 & 0.07 & \mycheckmark & \myxmark & 1.33 & 1.30 & 0.03 & \myxmark & \myxmark & \myxmark & \myxmark & \myxmark & \myxmark & \mycheckmark & \mycheckmark & \myxmark & \myxmark & \myxmark & 0.16 \\ %
\rowcolor{gray!20} \oursshort & 1.85 & 1.29 & 0.57 & \mycheckmark & \mycheckmark & 1.72 & 1.26 & 0.45 & \doublecheckmark & \mycheckmark & \mycheckmark & \doublecheckmark & \doublecheckmark & \doublecheckmark & \mycheckmark & \mycheckmark & \doublecheckmark & \mycheckmark & \doublecheckmark & 0.16 \\
\bottomrule
\end{tabular}
}
\caption{\textbf{Linguistic diversity}. We compare \oursshort to prior VLM-scaled datasets and \oursshortman, considering
the number of total, target, and anchor attributes/relationships (\textbf{all}, \textbf{tgt}, and \textbf{anc}); 
attributes such as \textbf{func}tion and \textbf{state}; 
relationships such as \textbf{far}, \textbf{arr}angements, \textbf{ord}inal and \textbf{comp}arisons; 
target references such as \textbf{gen}eric and coarse-grained (\textbf{CG}); 
anchor types such as \textbf{mul}tiple-object, \textbf{non}-object, and agent-based (\textbf{Agt}); 
language patterns such as the target not being the first noun (\textbf{NFN}) and \textbf{neg}ations; 
and the proportion of unique lexical bigrams (\textbf{2lex}). 
We use \mycheckmark if $\geq5\%$ of queries have the pattern and \doublecheckmark for $\geq20\%$ of queries. \oursshort has equal or better representation of all linguistic patterns over VLM-scaled datasets. See \cite{wang2024vigil3d} for details on each metric.
}
\label{tab:dataset-comparison}
\vspace{-3mm}
\end{table*}

\begin{table*}[tb]
\centering
\resizebox{\linewidth}{!}
{
\begin{tabular}{@{}rl lp{1in}r rr rrr rrrrrr @{}}
\toprule
& & \multicolumn{3}{c}{Data} & \multicolumn{2}{c}{\shortstack{\oursshortman\citeyearpar{wang2024vigil3d}}} & \multicolumn{3}{c}{Acc on ScanRefer~\citeyearpar{chen2020scanrefer}} & \multicolumn{6}{c}{F1 on Multi3DRefer~\citeyearpar{zhang2023multi3drefer}} \\
\cmidrule(){3-5} \cmidrule(l){6-7} \cmidrule(l){8-10} \cmidrule(l){11-16}
\# & Model & Pretrain & Finetune & Size & Acc & F1 & Uniq & Multi & All & ZT-D & ZT+D & ST-D & ST+D & MT & All \\
\midrule
1 & ZSVG3D & -- & -- & -- & 18.9 & 12.2 & 73.9 & 24.9 & 34.4 & 2.5 & 1.1 & 19.5 & 14.7 & 0.1 & 10.9 \\
2 & 3D-VisTA & ScanScribe & SR & 46K & 15.1 & 15.1 & 84.6 & \best{48.2}& 55.2 & 26.3 & 11.1 & 39.1 & 25.6 & 21.4 & 26.7 \\
3 & 3D-GRAND & 3D-GRAND & SR & 46K & 16.0 & 9.2 & 73.3 & 33.0 & 40.8 & 18.4 & 12.4 & 32.6 & 17.0 & 13.6 & 19.0 \\
4 & GPS & SceneVerse & SR & 46K & \best{24.1} & 14.7 & 88.1 & 46.4 & 54.5 & 0.0 & 0.0 & 40.6 & \best{28.1} & 31.9 & 29.1 \\
\midrule
5 & \oursmodel & ScanScribe & SR & 46K & 17.8 & 11.9 & 87.2 & 45.2 & 53.3 & 4.7 & 0.5 & 39.8 & 24.8 & 30.6 & 27.3 \\
6 & \oursmodel & ScanScribe & SR, M3D & 101K & 14.6 & 12.6 & 86.0 & 45.6 & 53.4 & \best{81.3} & 43.4 & 39.9 & 25.8 & 33.0 & 33.5 \\
7 & \oursmodel & ScanScribe & SR, M3D, SR3D & 185K & 19.2 & 13.6 & 88.6 & 46.9 & 55.0 & \second{79.5} & 44.2 & 41.2 & 26.6 & 32.4 & 33.9 \\
\midrule
\rowcolor{gray!20} 8 & \oursmodel & ScanScribe & \ourdataset & 105K & 18.6 & \second{15.5} & 80.9 & 37.0 & 45.5 & 25.6 & 14.6 & 30.7 & 18.6 & 26.7 & 23.1 \\
\rowcolor{gray!20} 9 & \oursmodel & ScanScribe & \ourdatasetlarge & 206K & \second{22.9} & \best{15.7} & \best{88.7} & \second{47.6} & \best{55.6} & \second{79.5} & \best{45.8} & \best{41.5} & \second{27.3} & \best{33.1} & \best{34.5} \\
\bottomrule
\end{tabular}
}
\caption{\textbf{Evaluation on 3DVG benchmarks}. ScanRefer (SR) is evaluated by accuracy, while Multi3DRefer (M3D) is evaluated by F1 (following prior work). Object proposals are based on GT. \oursmodel achieves competitive or better performance across all benchmarks.
}
\label{tab:model-evaluation-all-gt}
\end{table*}

  \subsection{Query Validation}

  We first investigate the following question: \textbf{\emph{does \oursshort generate more frequently valid 3DVG queries?}}
  We manually evaluate the validity based on 1) \emph{agreement}---the target(s), if any, should satisfy all of the constraints in the query---and 
  2) \emph{uniqueness}---no other objects in the scene should satisfy the constraints. 
  We compare \oursshort with three baselines for using VLMs to generate VG descriptions from images (\emph{Im2Query}), image-extracted captions (\emph{Cap2Query}), or scene graphs (\emph{SG2Query}) using GPT-4.1~\citep{achiam2023gpt}. 
  We also compare with 3D-GRAND~\citep{yang20243d}, which employs a similar scene graph-based method for 3DVG query generation.
  Since these baselines (see \cref{fig:teaser}) cover the methodologies used by other VLM-based generation methods, we re-implement them to isolate the key methodological differences (\eg by using similar prompt language).
  See \cref{appendix:generation-baselines} for details.
  
  For each baseline, we manually evaluate the validity of 50 queries of each target type given the point cloud, description, target(s), and an RGB video of the scene. For \oursshort and 3D-GRAND, we recruit external users to validate 200 queries (details in \cref{appendix:user-study}).
  We measure the proportion of correct descriptions and compute the 2-sigma Wald confidence interval for binomial proportions.

  \cref{tab:prompt-validation} shows that \textbf{\oursshort achieves higher validity than auto-scaled 3DVG generation methods}. Im2Query and Cap2Query often fail \emph{uniqueness} due to the limited scene visibility and lack of 3D understanding. 
  SG2Query suffers when generating queries with targets of common classes. It often hallucinates relationships, generates ambiguous referents (\eg ``next to the table and table''), or fails to contrast the target with distractors.
  While the validity rate reported for 3D-GRAND is 91.8\%, our study revealed only 38.3\% validity in a sample of their dataset due to failing the \emph{uniqueness} criterion by not contrasting the target with distractors. Users found that many 3D-GRAND queries are ambiguous, such as due to directional relationships, and most descriptions of common objects do not distinguish them from distractors (\eg two nightstands around a bed are described identically in a scene). Examples of invalid queries are provided in \cref{fig:failure-cases}.

  We identified several common failure cases for \oursshort. 
  Scene graph errors can lead to systemic errors in query generation, caused by imperfect point cloud reconstructions, VLM attribute prediction errors, and edge cases for geometric relationship definitions. 
  Attributes and relationships are also contextual. For instance, while there are intuitive definitions of \emph{near} and \emph{far}, in practice humans identify these based on relative distances of objects and the room size, making it challenging to encode them into a scene graph.
  We further evaluate our scene graph extraction in \cref{appendix:sg-evaluation} and provide example failure cases in \cref{fig:failure-cases-vigil3dpp}.
  Despite these challenges, we demonstrate that models can still learn from the complex spatial logic encoded in \oursshort.

  \subsection{Diversity Analysis}
  We next evaluate the following: \textbf{\emph{does \oursshort generate more diverse queries than existing methods?}}
  We compare against existing VLM-scaled datasets. \textbf{SceneVerse} \citep{jia2024sceneverse} leverages scene and object captions from VLMs to generate rephrased template-based referral descriptions. \textbf{3D-GRAND} \citep{yang20243d} generates grounding descriptions from image-extracted scene graph representations using a VLM. \textbf{3D-VisTA} \citep{zhu20233d} combines LLM-rephrased descriptions from existing ScanNet-based datasets with template-based descriptions extracted from 3RScan~\citep{Wald2019RIO} scene graph annotations. We also compare with \textbf{ViGiL3D}~\citep{wang2024vigil3d} as a benchmark for linguistic diversity in 3DVG.
  We use the analysis pipeline from \citet{wang2024vigil3d}, which measures the frequency of attribute and relationship types and language patterns for 3DVG queries. We analyze 1000 queries from each dataset except ViGiL3D, which has only 350 queries. 

  We show the results in \cref{tab:dataset-comparison}, finding that \textbf{\oursshort achieves better representation of linguistic patterns compared to existing methods}.
  The average number of attributes and relationships in \oursshort is closer to human descriptions than other scaled datasets while still capturing a range of constraint counts (\cref{fig:distribution-lengths}). \oursshort also represents more constraint types and target and anchor references, variability in sentence structure, and more cases of negation than LLM-scaled datasets. Although it does not outperform ViGiL3D, our method does not require manual annotations. 

  \subsection{Model Evaluation}
  We examine the training value of \oursshort: \textbf{\emph{when used for training, does \oursshort yield better model performance?}}
  We compare \oursmodel trained on \oursshort to comparable \emph{open-vocabulary} models trained on scaled 3DVG datasets, including \textbf{3D-VisTA}~\citep{zhu20233d}, \textbf{3D-GRAND}~\citep{yang20243d}, and \textbf{GPS}~\citep{jia2024sceneverse}. 
  We also compare to \textbf{ZSVG3D}~\citep{yuan2024visual}, a zero-shot method that uses an object localization module and an LLM to reason among candidate objects. 
  For each, we report the accuracy or F1 score, consistent with prior reporting. 
  We show the performance of each method when given ground truth object proposals in \cref{tab:model-evaluation-all-gt}.
  We report results using Mask3D~\citep{schult2023mask3d} predictions in \cref{appendix:model-experiments}.
   
  \cref{tab:model-evaluation-all-gt} shows that training on \oursshort (row 8) compared to ScanRefer and Multi3DRefer (row 6) yields better performance on a diverse benchmark like ViGiL3D for similar training set sizes (105K vs 101K). Although row 6 performs better on ScanRefer and Multi3DRefer, the training data is more in-distribution with those benchmarks.
  Additionally, training on \ourdatasetlarge (row 9) compared to the equivalent with SR3D~\citep{achlioptas2020referit3d} (row 7) shows that \ourdataset improves on simpler template-based descriptions all benchmarks, showing the benefit of its diversity and flexibility. Compared to prior methods (rows 1-4), our method matches or improves on all metrics.
  Lastly, training \oursmodel on Scanrefer (row 5) compared to 3D-VisTA (row 2) shows that this performance gain is from the data and not architectural changes.
  We thus find that \textbf{\oursmodel trained on \oursshort leads to better performance on diverse 3DVG benchmarks}.

\section{Conclusion}
  \label{sec:conclusion}

  Existing works for scaling 3DVG datasets fall short in generating valid, linguistically diverse descriptions, tending to generate verbose descriptions that inadequately distinguish the target from distractors.
  \oursshort, through cross-referencing and constraint sampling, enables the scalability of diverse grounding queries on a range of scene datasets, achieving higher validity and linguistic diversity than previous methods.
  Through scaling, we enable more robust training of 3DVG and foundation models, improving model performance on all benchmarks.
  Given the need for dataset scaling in 3D, we believe that this work can benefit the integration of 3D and language and encourage research to focus on diversifying the language on which models are trained. 
  Beyond 3DVG, the key idea of contrasting spatial entities through diverse language descriptions can apply to various vision-language problems, including VQA and embodied agents interacting with scenes.

\mypara{Limitations.}
  \oursshort relies on generating a comprehensive, high-quality scene graph. However, this is challenging due to errors in VLM-based attribute extraction; data quality, such as imperfect scene reconstructions; as well as the difficulty of converting human intuition about relationships into an explicit graph. \oursshort also does not support certain relationship types such as allocentric directional relationships (\eg objects with a canonical front), emphasizing the challenge of building a complete scene graph. This work also highlights the current deficiencies of VLMs; they can generate reasonable-sounding descriptions, but they fall short at precise spatial reasoning. When provided with many parameters in the prompt, they have difficulty resolving the inputs into a coherent output, such as when providing a long list of constraints to rephrase. Further improvements in instruction-following for VLMs and capturing the diversity of potential outputs are thus needed.

\section*{Acknowledgement}
\label{sec:acknowledgement}
This work was funded in part by a CIFAR AI Chair and the NSERC Discovery Grant, and enabled by support from the Digital Research Alliance of Canada and a CFI/BCKDF JELF.
We thank Hou In Ivan Tam, Hou In Derek Pun, Denys Iliash, Weikun Peng, Tristan Engst, and Xingguang Yan for the helpful discussions and feedback.

{
    \small
    \bibliography{main}
    \bibliographystyle{styles/colm/colm2026_conference}
}

\vfill\eject  %
\appendix

\section*{Appendices}

We provide additional information about baselines and prior work (\appref{appendix:related-works}) as well as details about \oursshort (scene graph extraction in \appref{appendix:sg-extraction} and constraint solving in \appref{appendix:constraint-solver}), the test dataset (\appref{appendix:dataset}), and
\oursmodel details (\appref{appendix:model}). Documentation of the scientific artifacts used in this work can be found in \appref{appendix:attribution}.

\section{Comparison to Prior Work}
  \label{appendix:related-works}
  \begin{table*}[bp]
\centering
\resizebox{\linewidth}{!}
{
\begin{tabular}{@{}lp{3in} cccccc}
\toprule
Dataset                                         & Scene Datasets & exist        & struct       & cap          & img          & tmpl         & sg            \\
\midrule
3D-LLM~\citeyearpar{hong20233d}                 & ScanNet, HM3D   & \mycheckmark & \mycheckmark & \myxmark     & \mycheckmark & \myxmark     & \myxmark     \\
M3DBench~\citeyearpar{li2023m3dbench}           & ScanNet         & \mycheckmark & \mycheckmark & \mycheckmark & \myxmark     & \myxmark     & \myxmark     \\
ScanScribe~\citeyearpar{zhu20233d}              & ScanNet, 3RScan & \mycheckmark & \myxmark     & \myxmark     & \myxmark     & \mycheckmark & GT           \\
LEO-Align~\citeyearpar{huang2023embodied}             & ScanNet, 3RScan & \mycheckmark & \myxmark     & \myxmark     & \myxmark     & \mycheckmark & 3DSSG        \\
EmbodiedScan~\citeyearpar{wang2024embodiedscan} & ScanNet, 3RScan, Matterport3D & \myxmark     & \myxmark     & \myxmark     & \myxmark     & \mycheckmark & geo          \\
MMScan~\citeyearpar{lyu2024mmscan}              & ScanNet, 3RScan, Matterport3D & \myxmark     & \myxmark     & \mycheckmark & \myxmark     & \mycheckmark & \myxmark     \\
ScanReason~\citeyearpar{zhu2024empowering}      & ScanNet, 3RScan, Matterport3D & \myxmark     & \mycheckmark & \myxmark     & \myxmark     & \myxmark     & \myxmark     \\
SceneVerse~\citeyearpar{jia2024sceneverse}      & ScanNet, 3RScan, HM3D, MultiScan, ARKitScenes, ProcTHOR, Structured3D & \mycheckmark & \myxmark     & \mycheckmark & \myxmark     & \mycheckmark & VLM,geo      \\
3D-GRAND~\citeyearpar{yang20243d}               & 3D-FRONT, Structured3D & \myxmark     & \myxmark     & \myxmark     & \myxmark     & \mycheckmark & VLM          \\
\rowcolor{gray!20} \oursshort                   & ScanNet, 3RScan, MultiScan & \mycheckmark & \myxmark     & \myxmark     & \myxmark     & \mycheckmark & VLM,geo      \\
\bottomrule
\end{tabular}
}
\caption{\textbf{Dataset overview}. Overview of 3D visual grounding datasets, the scenes used in the datasets, and how the VG descriptions are generated. We compare popular and recent human-annotated datasets with recent work that uses MLLMs to generate descriptions.
}
\label{tab:all-dataset-overview-general}
\end{table*}

  \begin{table*}[thbp]
\centering
\small
\begin{tabular}{@{}l p{0.75\textwidth} @{}}
\toprule
Dataset & Example Description \\
\midrule
ScanRefer~\citeyearpar{chen2020scanrefer} & {\small There is a black counter top to the left of the fridge. It has a stainless steel sink on it.} \\
Beacon3D~\citeyearpar{huang2025unveiling} & {\small Gray chair next to the corner table}\\
\midrule
M3DBench~\citeyearpar{li2023m3dbench} & {\small There is a black cylindrical trash can, which is
located to the right of the kitchen counter and in
front of the printer. Please give me the
coordinates of its center and the length, width, and height of the bounding box.} \\
ScanScribe~\citeyearpar{zhu20233d} & {\small The chair is behind the desk, on the left side of the circular black table, to the right of the rectangular shelf, and to the left of the other chair, bag, trash can, and backpack.} \\
LEO-Align~\citeyearpar{huang2023embodied} & {\small The rectangular wooden table is elegantly positioned on the floor, boasting a rich brown hue. It stands at a considerable height and is accompanied by several surrounding chairs. To the left of the table, a tall chair catches the eye, while adjacent to it is another rigid and tall chair. Notably, there is also another tall chair nearby.} \\
EmbodiedScan~\citeyearpar{wang2024embodiedscan} & {\small find the bag that is closer to the bathtub} \\
SceneVerse~\citeyearpar{jia2024sceneverse} & {\small The couch is situated to the right of the computer tower.} \\
3D-GRAND~\citeyearpar{yang20243d} & {\small This refrigerator is a muted silver, presenting a sleek and modern look with its brushed metal finish. The refrigerator is positioned close to one of the dining chairs, near to another dining chair, and far from the loveseat sofa.} \\

\rowcolor{gray!20} \oursshort  & {\small From your position beside the table and facing the wall, you can see containers resting on an object close to the oven.} \\
\bottomrule
\end{tabular}
\caption{\textbf{Dataset example descriptions}. We provide example characteristic descriptions from human-annotated and MLLM-scaled VG datasets. In aggregate, \oursshort demonstrates more linguistic diversity and complexity than existing benchmarks.
}
\label{tab:all-dataset-overview-example}
\end{table*}

  We provide additional details and example descriptions on different MLLM-scaled 3DVG query generation methods in \Cref{tab:all-dataset-overview-general,tab:all-dataset-overview-example}. 
  While many methods use scene graphs to generate prompts, ScanScribe~\citep{zhu20233d} relies on ground truth annotations~\citep{wald2020learning} and thus does not scale to  datasets without scene-graph annotations. LEO-Align~\citep{huang2023embodied} uses 3DSSG~\citep{wald2020learning} trained on those annotations and is thus limited in the types of relations it is trained on.
  
  Additionally, we find that most scaled 3DVG datasets are either overly simple or verbose. For example, EmbodiedScan~\citep{wang2024embodiedscan} and SceneVerse~\citep{jia2024sceneverse} rely on template-based methods and thus can only really capture one relationship in a given prompt. On the other hand, the other methods in \cref{tab:all-dataset-overview-example} include a large number of attributes and relationships and typically repeat linguistic patterns (\eg starting the sentence with the target immediately).

\subsection{Description Generation Baselines}
  \label{appendix:generation-baselines}
  \begin{listing*}
  \centering
  \begin{lstlisting}[language=Python,frame=single,breaklines=true]
"""
Given an object and a list of views of the object, please provide a visual grounding description, in the style of ScanRefer
describing the object in the image. The object of interest is marked in the images with a red bounding box. The description should uniquely identify the object and differentiate it from others in the scene by describing attributes of objects and spatial, geometric, or functional relationships between objects in the scene, and it should be clear what object the description refers to. The description should also be consistent with every image view of the object.

Example descriptions include:
1. This is a cardboard box against the wall and right next to two doors. Of such boxes, it is the one not next to an armchair.
2. On the brown, wooden table is a small, rectangular projector.
3. This is the larger of the two toolboxes near the piano.
4. Find the object with wheels.
5. When facing the radiator, the metal rail is directly on your left.
6. To the left of the sink is a set of paper towel rolls.
7. This circular object is supported by a white mini fridge.
8. Underneath the sink is a cabinet for storing cleaning supplies.
9. Locate the tallest appliance in the kitchen.
10. There is a row of black rolling chairs along the wall. Pick the one furthest to the left when facing them.

You should return {num_candidates} candidate descriptions, separated by "|", and no other text. The descriptions should vary in content and style.
"""
  \end{lstlisting}
  \caption{\textbf{Im2Query Prompt.} We provide a list of annotated multiple views and the target name and ask the VLM to generate several candidate grounding descriptions.
  }
  \label{prompt:im2prompt}
\end{listing*}

  \begin{listing*}
  \centering
  \begin{lstlisting}[language=Python,frame=single,breaklines=true]
CAPTION_PROMPT = """
Given an object and a list of views of the object, please provide a caption description of all of the attributes of the object and relationships with objects around them. The attributes should include any of colors, size, shape, material, texture, style, text on the object, the state of the object (e.g. open, closed), or descriptions of parts of the objects. Relationships should include a description of any objects around the object of interest, such as objects nearby or far away, above or below, or other such relationships.

You should return only the object description with no other text.
"""
GROUNDING_PROMPT = """
Given a caption of an object, generate a grounding description of the object. The description should uniquely identify the object and differentiate it from others in the scene by describing attributes of objects and spatial, geometric, or functional relationships between objects in the scene, and it should be clear what object the description refers to.

Example descriptions include:
1. This is a cardboard box against the wall and right next to two doors. Of such boxes, it is the one not next to an armchair.
2. On the brown, wooden table is a small, rectangular projector.
3. This is the larger of the two toolboxes near the piano.
4. Find the object with wheels.
5. When facing the radiator, the metal rail is directly on your left.
6. To the left of the sink is a set of paper towel rolls.
7. This circular object is supported by a white mini fridge.
8. Underneath the sink is a cabinet for storing cleaning supplies.
9. Locate the tallest appliance in the kitchen.
10. There is a row of black rolling chairs along the wall. Pick the one furthest to the left when facing them.

You should return {num_candidates} candidate descriptions, separated by "|", and no other text. The descriptions should vary in content and style.
"""
  \end{lstlisting}
  \caption{\textbf{Cap2Query Prompt.} We provide a list of annotated multiple views for each object and ask the VLM to generate captions for each. We then provide the caption of the sampled target and ask the VLM to generate a corresponding grounding description.
  }
  \label{prompt:cap2prompt}
\end{listing*}

  \begin{listing*}
  \centering
  \begin{lstlisting}[language=Python,frame=single,breaklines=true]
SG_SINGLE_PROMPT = """
Given a scene graph describing the objects, attributes, and relationships of a scene and a specified target ID, generate a grounding description of one of the objects. The description should uniquely identify the object and differentiate it from others in the scene by describing attributes of objects and spatial, geometric, or functional relationships between objects in the scene which are true of the target or other referenced objects and not of others in the scene. It should be clear what object the description refers to. The candidate descriptions must include the id of each mentioned object, when known, to distinguish different referents using the syntax "object_name<id:X>", where X is the ID number. If a noun refers to multiple objects, specify this with commas as <id:X,Y,Z>. In your final output description, please use the corresponding ID tags for all references to objects, including coreferences.
        
Example descriptions include:
- This<id:3> is a cardboard box<id:3> against the wall<id:15> and right next to two doors<id:10>. Of such boxes<id:3,7>, it is the one<id:3> not next to an armchair<id:12>.
- On the brown, wooden table<id:13> is a small, rectangular projector<id:1>.
- This<id:24> is the larger of the two toolboxes<id:24,40,41> near the piano<id:5>.
- Find the object<id:26> with wheels.
- When facing the radiator<id:20>, the metal rail<id:8> is directly on your left.
- To the left of the sink<id:4> is a set of paper towel rolls<id:101>.
- This circular object<id:18> is supported by a white mini fridge<id:17>.
- Underneath the sink<id:23> is a cabinet<id:31> for storing cleaning supplies.
- Locate the tallest appliance<id:25> in the kitchen.
- There is a row of black rolling chairs<id:3,4,5,6> along the wall<id:29>. Pick the one<id:5> furthest to the left when facing them<id:3,4,5,6>.

If referencing an <AGENT> in the scene graph, refer to it as the implied audience without an ID, e.g. "when standing in front of the couch", "to your left", or "behind you".

Use provided reasons for previous failures if any, to generate better descriptions which avoid such failures.

You should return {num_candidates} candidate descriptions, separated by "|", and no other text. The descriptions should vary in content and style. All descriptions should be grammatically correct and should not include a '-' at the beginning of the prompt.
"""
SG_ZERO_PROMPT = """
Given a scene graph describing the objects, attributes, and relationships of a scene and a candidate description of an object in the scene, modify the description such that it no longer describes any of the objects in the scene. You should try to preserve as much of the sentence structure as possible but change attributes, relationships, or object labels to make the description no longer applicable. Modifications can include but are not limited to
- Swapping a single attribute to one that no longer applies (e.g. "This is the rectangular table" to "This is the circular table".)
- Swapping a single relationship to one that no longer applies (e.g. "This is the table next to the chair" to "This is the table on the chair".)
- Swapping two object labels (e.g. "This is the table to the left of the chair" to "This is the chair to the left of the table".)
- Swapping a single object label to one that no longer applies (e.g. "This is the wooden wardrobe for holding clothes" to "This is the wooden water bottle for holding clothes".)
- Applying negation (e.g. "Look for the wooden table" to "Look for the table not made of wood")

The new description cannot be true of a different object in the scene.

You should return a single candidate modified description. The description should be grammatically correct, and no other text should be included.
"""
SG_MULTI_PROMPT = """
Given a scene graph describing the objects, attributes, and relationships of a scene and specified target IDs, generate a grounding description of the target objects. The description should uniquely identify all of the target objects and differentiate it from others in the scene by describing attributes of objects and spatial, geometric, or functional relationships between objects in the scene which are true of the targets or other referenced objects and not of others in the scene. It should be clear what object the description refers to. The candidate descriptions must include the id of each possible object for each noun phrase, when known, to distinguish different referents using the syntax "object_name<id:X>", where X is the ID number. If a noun can refer to multiple possible objects, specify this with commas as <id:X,Y,Z>. In your final output description, please use the corresponding ID tags for all references to objects, including coreferences.
...
"""
  \end{lstlisting}
  \caption{\textbf{SG2Query Prompt.} We provide the scene graph in text form along with the desired target objects and ask the VLM to generate several candidate grounding descriptions. For zero-target prompts, we follow up by prompting the VLM to modify a single-target prompt into a zero-target one. For the multi-target prompt, the rest of the prompt description is similar to the single-target prompt, so it is omitted for brevity.
  }
  \label{prompt:sg2prompt}
\end{listing*}

  We compare our method against several baseline approaches for generating grounding descriptions and show that a simple approach to generating queries would not yield sufficiently diverse or correct queries. For fair comparison, we use GPT-4.1~\citep{achiam2023gpt} as the MLLM in all methods, and prompts used for each baseline in \cref{prompt:im2prompt,prompt:cap2prompt,prompt:sg2prompt}. 
  
  \textbf{\textit{Im2Query}} generates queries by generating grounding descriptions from multiview images of a target object.
  For each query, we sample a target object from the scene. We sample $n_{\text{view}} = 3$ image views, using an identical method to \oursshort. The target name, the object ID, and the image views are then passed to a VLM to generate a candidate description. See \cref{prompt:im2prompt} for the prompt we used.
  
  \textbf{\textit{Cap2Query}} is implemented similarly to \textit{Im2Query}, but instead of generating grounding descriptions directly from images, intermediate captions are generated for each object first. The captions include both attributes and relationships. Grounding descriptions are generated using the provided target caption. See \cref{prompt:cap2prompt} for the prompts for generating captions and grounding queries.
  For Im2Query and Cap2Query, we only generate single-target queries due to the large input required to provide the necessary context for the rest of the scene to also generate zero- and multi-target queries.

  \textbf{\textit{SG2Query}} relies on the same scene graph input to \oursshort but passes the entire scene graph directly to the VLM prompt along with the sampled target objects to generate candidate descriptions, thus serving as the closest baseline to \oursshort without the constraint sampler. Zero-target queries are generated by first generating a description for a single target and then prompting an LLM to modify the query to make it invalid. Multi-target queries are generated by sampling a label and a subset of target objects with that label. All three prompts are shown in \cref{prompt:sg2prompt}.

\clearpage
\section{Scene Graph Extraction}
  \label{appendix:sg-extraction}
    \begin{figure*}[t]
  \centering
  \includegraphics[trim={0 0 0 4px},clip,width=\linewidth]{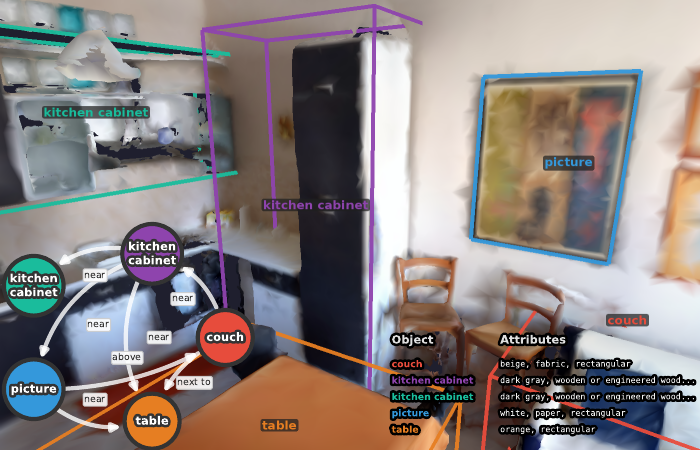}
  \caption{\textbf{Scene graph visualization.} We visualize a subgraph of the scene graph for the following scene, including a sample of the attributes and relationships extracted for each.
  }
  \label{fig:sg-visualization}
\end{figure*}

  \begin{listing*}[bp]
  \centering
  \begin{lstlisting}[language=Python,frame=single,breaklines=true]
"""
Given a list of object classes, decide which other classes also apply. You should particularly consider the following categories:
* "object" - any entity which is not part of the architecture and can be moved in the scene (e.g., boxes, chairs, tables, small objects). For instance, the 
* "thing" - any entity in the scene
* "device" - any small-scale entity which requires a power supply to perform a function (e.g., laptop, monitor, phone, lamp)
* "appliance" - a device designed to perform a specific function for household use (e.g., toaster, oven, fridge)
* "storage" - any object which is used for storing other objects (e.g., shelf, cabinet, box)
* "container" - any movable object which stores other objects in an enclosed space (e.g., box, jar, case)
* "furniture" - any large object used to make a space suitable for living or working (e.g., tables, chairs, shelves)

For any categories you use, make sure to consider if they might apply to other object classes. If none of the other proposed object categories apply, leave the list empty.

Provide your response as a JSON of the following form:
{
    "classes": [
        {
            "label": "class_1",
            "additional_labels": [<list of additional categories>]
        },
        {
            "label": "class_2",
            "additional_labels": [<list of additional categories>]
        },
        ...
    ]
}

Make sure to include each object class and list as a dictionary under the "tags" key.
"""
  \end{lstlisting}
  \caption{\textbf{Object Tagging Prompt.} We provide the full list of object labels to an LLM and ask it to return a list of additional potential labels (\eg ``object,'' ``appliance,'' or ``device'') for each class.
  }
  \label{prompt:object-tagging}
\end{listing*}

  \begin{listing*}[t]
  \centering
  \begin{lstlisting}[language=Python,frame=single,breaklines=true]
"""
You are given images of two objects (marked by red outlines) and their proposed object labels. Identify if the objects in the images are identical in attributes and appearance. Return the result as a JSON of the following form:
{
    "is_same": true or false
}
"""
  \end{lstlisting}
  \caption{\textbf{Object similarity prompt.} We prompt the VLM to validate whether the two objects shown in the images are similar, \ie should be assigned the same attributes by the prompt.
  }
  \label{prompt:similarity}
\end{listing*}

  \begin{listing*}[t]
  \centering
  \begin{lstlisting}[language=Python,frame=single,breaklines=true]
"""
### INSTRUCTIONS ###
You are an assistant tasked with extracting the attributes of an object from an image.
Given the object category name and image views of an object, please identify the attributes of the object which you can discern from any of the images based on the categories.
The object of interest should be marked with the ID label shown in the image and the red outline around the object.

Requirements:
- Attributes should not be conflicting across views. 
- Only include those attributes which you are highly confident in.
- Provide only the JSON output based on the extracted information.

The JSON format is as follows, with the provided definitions and examples:
{
    "color": list of attributes describing the main colors of the object (e.g., "red", "blue", "green"),
    "size": list of attributes describing how small or large the object (e.g., "small", "large", "short", "tall"),
    "shape": list of attributes representing the shape of the object (e.g., "round", "square", "circular", "rectangular", "cylindrical"),
    "material": list of attributes describing the materials likely comprising the object (e.g., "wooden", "metal", "plastic"),
    "texture": list of attributes describing how the object would feel if touched (e.g., "smooth", "rough", "bumpy", "squishy", "comfortable"),
    "function": list of attributes describing what the object does (e.g., "for sitting on", "for eating food on"),
    "style": list of attributes describing the aesthetic style of the object (e.g., "modern", "vintage", "retro"),
    "text_label": list of attributes describing any text on the object, not including the ID mark of the object. For text labels, explicitly preface the attribute with the word "labeled" and the text in quotes (e.g., "labeled 'exit'"),
    "state": list of attributes describing a changeable state of the object (e.g., "open", "closed", "folded"). If the object does not have a changeable state, return an empty list.
}
All attribute types must be present in the JSON. Please use an empty list if no attributes are described for a particular attribute type. Any relationships to other objects should not be included. 
You should return a JSON and only the JSON. Avoid including other text or explanations.
"""
  \end{lstlisting}
  \caption{\textbf{Attribute Extraction Prompt.} We provide multi-view images of an object to a VLM and ask it to return a JSON of attributes to assign to the object, organized by attribute type.
  }
  \label{prompt:attribute-extraction}
\end{listing*}

  \begin{listing*}[t]
  \centering
  \begin{lstlisting}[language=Python,frame=single,breaklines=true]
"""
### INSTRUCTIONS ###
You are an agent trying to identify the attributes of an object to populate a scene graph. Given the object category name and image views of an object, identify any of the attributes that accurately describe the red-outlined object in the images. Please only include the high-confidence attributes. Provide your response as a JSON of the following form:
{ 
    "attributes": [<list of applicable attributes>] 
}
"""
  \end{lstlisting}
  \caption{\textbf{Attribute Cross-Referencing Prompt.} We provide multi-view images of an object to a VLM as well as a list of attributes and ask it to return a JSON of the additional attributes to assign to the object.
  }
  \label{prompt:attribute-cross-reference}
\end{listing*}

  \begin{listing*}[t]
  \centering
  \begin{lstlisting}[language=Python,frame=single,breaklines=true]
"""
### INSTRUCTIONS ###
You are an agent in a scene trying to identify the support relationship between two objects based on an image. You should pick one of the following relationships:
1. "on" (inverse: "supports"): if the subject object is physically resting on top of the recipient object (e.g. a book on a table, a bottle on a shelf).
2. "in" (inverse: "contains"): if the subject object is physically inside, or embedded inside, the recipient object (e.g. a book in a box, a cup in a cupboard, door in a wall).
3. "embedded in" (inverse: "embeds"): if the subject object is partially inside the recipient object, such that part of the subject is accessible at the surface of the recipient (e.g. a stove embedded in the counter, a sink embedded in the counter).
4. "leaning on" (inverse: "supports"): if the subject object is leaning against the recipient object, such that if you were to remove the recipient, the subject would obviously fall. Horizontal contact of one object against another on its own does not count as leaning (e.g. a ladder leaning on a wall, an instrument case leaning against a table).
5. "hanging on" (inverse: "supports"): if the subject object is hanging from the recipient object (e.g. a towel hanging on a rack, a lamp hanging from the ceiling, a picture hanging on the wall).
5. "none" (inverse: "none"): if none of the above relationships apply.

# How to Determine the Relationship
You will be provided the IDs and labels of the objects. The objects of interest should be marked with blue (subject) and orange (recipient) outlines around the objects, and ignore the highlighted red area. You should determine the appropriate relationship between the subject and recipient and the respective inverse (recipient to subject) based on two factors:
1. How are the objects typically described? (e.g. if a book is supported by a shelf, the book is usually described as "on" the shelf and not "in" the shelf)
2. Are the objects in the scene actually exhibiting the relationship? (e.g. if the book is actually only next to the table, then the relationship is "none")

# Output Format
The output should be a JSON object with the following format:
{
    "subject_to_recipient": one of ["supports", "on", "contains", "in", "leaning on", "hanging on", "none"] that indicates whether subject is supporting or supported by the recipient,
    "recipient_to_subject": opposite relationship type; one of ["supports", "on", "contains", "in", "leaning on", "hanging on", "none"],
    "reason": explanation for why the relationship was chosen
}
"""
  \end{lstlisting}
  \caption{\textbf{Support Relationship Prompt.} We provide an image view of the candidate objects to a VLM and ask it to predict which support relationship, if any, is applicable to the pair of objects.
  }
  \label{prompt:support-relationships}
\end{listing*}

In this section, we provide details of how we extract a scene graph for an 3D scene (\cref{appendix:sg-method}), prior works on scene graph extraction (\cref{appendix:sg-prior}), and an evaluation of our extracted scene graphs (\cref{appendix:sg-evaluation}).

  \subsection{Scene Graph Extraction Details}
  \label{appendix:sg-method}
  Our scene graph extraction consists of first extracting the room layout (\cref{appendix:sg-method-layout}), then identifying the objects and attributes (\cref{appendix:sg-method-attributes}) and relationships between objects (\cref{appendix:sg-method-rel-independent,appendix:sg-method-rel-dependent}). 
  A visualization of a subset of a scene graph is shown in \cref{fig:sg-visualization}.

  \subsubsection{Scene Layout}
  \label{appendix:sg-method-layout}

    \begin{figure*}[t]
  \centering
  \includegraphics[trim={0 0 0 4px},clip,width=\linewidth]{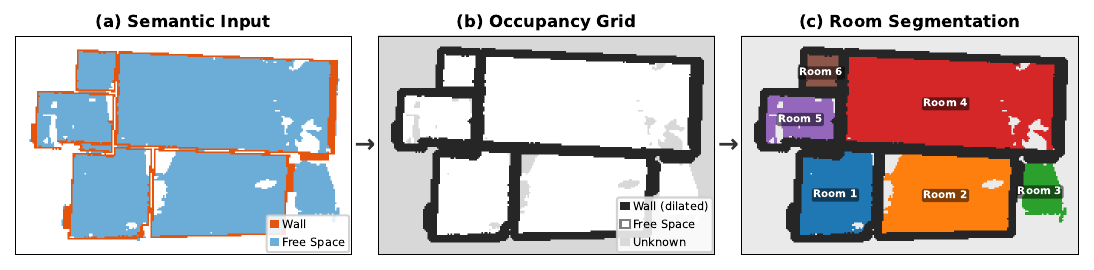}
  \caption{\textbf{Layout estimation pipeline.} 
  (a) Semantic annotations from the 3D mesh are projected onto the XY plane to identify walls and free space (floor and objects). 
  (b) A 2D occupancy grid is constructed at 5 cm resolution, where walls are dilated using morphological operations to close gaps at doorways and thin partitions. 
  (c) Connected component labeling on the free space regions segments the scene into individual rooms. Regions smaller than a minimum area threshold are filtered out as noise.
  }
  \label{fig:layout-estimation-pipeline}
\end{figure*}

  Room layout estimation (also referred to as structured indoor modeling or floorplan reconstruction)---involving predicting structural components, such as rooms, doors, and windows, from an input point cloud---is important for grounding. In multi-room scenes, meaningful spatial relationships are typically between objects in the same room, and grounding descriptions often refer to intra-room locations such as the \emph{corner} or the \emph{center} of the room. Without room data, certain relationships may be incorrect or impractical to solve (\eg objects in different rooms but on opposite sides of the same wall may be incorrectly predicted as \emph{near} each other).

  Prior work has proposed a variety of layout estimation methods~\citep{mura2014automatic, ochmann2016automatic, yue2023connecting, xu2024fri, ye2025floorsam, SpatialLM}. For our grounding task, however, a lightweight heuristic is sufficient, as we only need to know the room each object is in and approximate corner and center positions, neither of which requires precise floorplan geometry. Furthermore, since we already have a segmentation of structural classes (\textit{wall}, \textit{floor}, \textit{ceiling}), we can derive this information directly from the segmentation without adding a separate module to predict structural elements.

  We construct the scene layout by taking an annotated 3D mesh, and creating a 2D grid that identifies the walls and free space, and then segmenting out the rooms and determine semantic room labels (\cref{fig:layout-estimation-pipeline}).
  We build a 2D occupancy grid by vertically projecting the segmented point cloud, marking cells as free space where \textit{floor} or object points project, and as wall space where \textit{wall} points project. We apply binary dilation to reduce noise, extract connected components as candidate rooms, and filter by size. For each room, we fit a minimum-area rectangle to the convex hull of its boundary cells to obtain corners and a center. To assign a semantic label (\eg dining room), we render a top-down view and query a VLM to predict the label. An object is assigned to a room if the majority of its bounding-box corners lie within the room's free cells.
  
  \subsubsection{Object labels and Attributes}
  \label{appendix:sg-method-attributes}

  To predict attributes, we extract multi-view images for each object and prompt the VLM in multiple rounds: 1) identify similar objects, 2) predict attributes of a representative object per similarity group, and 3) augment the attributes for each object group with attributes assigned to other objects.

  \mypara{Image rendering.} To select images for each object, we use existing RGB-D images along with 2D segmentation maps per image, where provided, to select optimal views of each object. For each object, we first filter images by determining all frames in the top 25th percentile of area (number of pixels with corresponding object ID). We then filter for frames where the object is not at the edge (\ie not partially visible), and then we sample uniformly in time from the remaining frames. We process each frame before passing it to the VLM by cropping the object (with a relative margin of 0.3), resizing the frame to a maximum dimension of 384 pixels, and then masking the image to highlight the object of interest, as inspired by \citet{yang2023set}. 
  In cases where segmentation maps do not exist, they can be obtained by rasterizing the point cloud instance labels onto an image given the camera pose. Additionally, if RGB-D frames are not available, they can be rendered from the mesh using similar rasterization techniques.

  \begin{figure}[bp]
  \centering
  \includegraphics[trim={0 0 0 5px},clip,width=\linewidth]{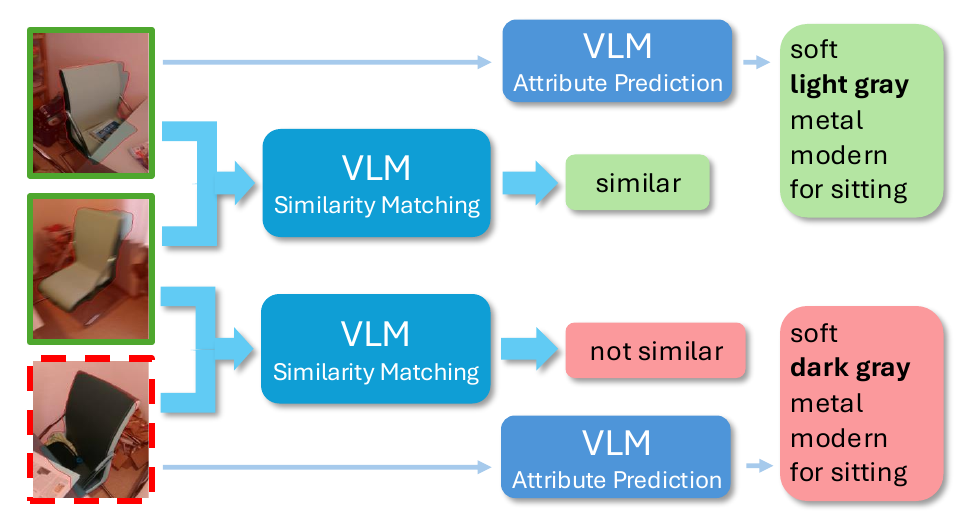}
\captionof{figure}{
  \textbf{Identifying similar objects.} 
  To ensure consistent assignment of attributes to objects, we first identify all groups of objects that have the same attributes, thus ensuring that attributes are assigned consistently among identical objects. 1) The two chairs share all of the same attributes, including color, whereas 2) the dark chair has a different color and thus is assigned attributes separately.
}
  \label{fig:cross-reference}
\end{figure}

  \mypara{Object category.} To enrich the list of possible category names for each objects in the scene, we provide the list of objects labels to an LLM and prompt it for additional labels for each class (\cref{prompt:object-tagging}).

  \mypara{Similarity matching.} Within each object class, we define two objects to be similar if they have the same attributes across \emph{all} types (\eg identical dining chairs). An illustration of this definition is shown in \cref{fig:cross-reference}. We cluster objects into similar groups using a greedy clustering algorithm. Namely, we query the VLM for similarity between each object and a representative object of each similarity group with the same object label, adding the object to the group if they match according to the VLM or creating a new group otherwise. The similarity matching prompt is given in \cref{prompt:similarity}.

  \mypara{Attribute prediction.}
  We first predict attributes by querying a VLM with images of the representative object per group to provide a list of attributes within each of the considered types: color, size, shape, material, texture, function, style, text label, and state (\eg open or closed). We pool all the attributes into a list and subsequently query the VLM with the full list of assigned attributes to determine which other attributes not initially assigned also apply to each representative object. The returned list of attributes is then applied to every object in each group of similar objects. The prompts for predicting attributes are provided in \cref{prompt:attribute-extraction} and \cref{prompt:attribute-cross-reference}.

  \subsubsection{Query-Independent Relationships}
  \label{appendix:sg-method-rel-independent}

  During scene graph extraction, we predict \emph{query-independent} relationships that are true regardless of the viewpoint or the subset of objects being considered. We support the following relationship types and categorize them according to \cite{wang2024vigil3d}:

  \mypara{Near.} Object \texttt{A} is \emph{next to} object \texttt{B} if it is within $d_{\text{next}}$ meters and \emph{near} if within $d_{\text{near}}$ meters. We set thresholds at $d_{\text{next}} = \SI{0.46}{\metre}$ and $t_{\text{near}} = \SI{1.22}{\metre}$, inspired by the study of proxemics~\citep{DANESI2006241}, which proposes general ranges of spaces for human interpersonal interactions. Furthermore, for small rooms, we clip the thresholds to $25\%$ of the diagonal of the room, \ie $\frac{1}{4}\sqrt{h^2+w^2}$.

  \mypara{Far.} Object \texttt{A} is \emph{far from} object \texttt{B} if they are in the same room and greater than $d_{\text{far}}$ meters apart. Following~\cite{DANESI2006241}, we set $t_{\text{far}} = \SI{2.13}{\metre}$.

  \mypara{In.} Object \texttt{A} is \emph{in} a room if the majority of its bounding box coordinates coincide in the occupancy grid of a particular room. \texttt{A} is considered in a corner of the room if the object is in the room and its center is within $t_{\text{corner}}$ meters of the corner. Lastly, \texttt{A} is considered in the center of a room if it is within the room and its center is within $d_{\text{center}}$ meters of the center of the room.

  \mypara{Higher/lower.} Object \texttt{A} is \emph{higher} than object \texttt{B} if the bottom of \texttt{A} is higher than the bottom of \texttt{B} (with a margin) and the top of \texttt{A} is higher than the top of \texttt{B}:
  \begin{align}
      \min{A_z} &> \min{B_z} + d_{\text{higher}} \\
      \max{A_z} &> \max{B_z} + d_{\text{higher}}
  \end{align}

  \mypara{Over/under.} Object \texttt{A} is \emph{over} object \texttt{B} if the vertical projection of \texttt{A} overlaps with \texttt{B} and \texttt{A} is completely higher than \texttt{B} within the overlapping region (with a margin). Letting $A^o \subseteq A$
  and $B^o \subseteq B$ refer to the subset of points of each that are in the overlapping projections onto the $xy$-plane, we formally define \emph{over} as follows:
  \begin{align}
      \text{ConvexHull}(A_{xy}) &\cap \text{ConvexHull}(B_{xy}) \neq \emptyset \\
      \min{(A_z^o)} &> \max{B_z^o} + t_{\text{over}}
  \end{align}
  
  \mypara{Above/below.} Object \texttt{A} is \emph{above} object \texttt{B} if \texttt{A} is completely higher than \texttt{B} (with a margin) and the vertical projection of \texttt{A} overlaps with \texttt{B}:
  \begin{align}
      \text{ConvexHull}(A_{xy}) &\cap \text{ConvexHull}(B_{xy}) \neq \emptyset \\     \min{A_z} &> \max{B_z} + t_{\text{above}}
  \end{align}
  Thus, if \texttt{A} is above \texttt{B}, then \texttt{A} is also higher than and over \texttt{B}.

  \mypara{Support.} Object \texttt{A} can be supported by \texttt{B} by being \emph{in}, \emph{on}, \emph{leaning on}, or \emph{hanging on}. Applying geometric methods to supporting objects represented by a point cloud is difficult due to the noise and sampled nature of point cloud reconstructions and the semantic nuances of relationship types. Thus, we use VLMs to identify the relationship between potential supporting objects. Given two objects whose point clouds are geometrically within $d_{support}$ meters away, we render an image with both objects in view, using a process similar to attribute extraction, and query a VLM to predict which relationship is true, since the choice of relationship depends on the semantics of the objects. We include our prompt to determine the relevant support relationship in \cref{prompt:support-relationships}.

  \subsubsection{Query-Dependent Relationships}
  \label{appendix:sg-method-rel-dependent}

  During constraint sampling, we sample \emph{query-dependent} relationships, which vary based on the viewpoint and objects being considered. Viewpoint-based relationships assume an agent in the scene from which certain relationships are described. Other relationships, such as comparisons (\eg taller, closest), may vary depending on the objects being considered (\eg the closest of all \emph{red} chairs to the fireplace).

  \mypara{Directional relationships.} For simplicity, we first sample a pose for an agent in the scene by placing the agent near an object or in a particular region of a room (\ie corner or center).
  We select one of these modes at random equally and then sample an object or region with equal chance. Only objects within the same room as the target are considered.
  For a sampled pose, we identify directional relationships radially within the $xy$-plane. For each object $o\in O$, we compute the minimum and maximum radial distances $r_{o\min}$ and $r_{o\max}$ of the projected 2D bounding box as well as the angular bounds $\theta_{o\min}$ and $\theta_{o\max}$. We also compute the distances and angles to the centers of each box as $r_{o\,\text{center}}$ and $\theta_{o\,\text{center}}$, respectively.
  
  For objects $o_i, o_j\in O$, $o_i$ is in front of $o_j$ if 
  
  \begin{gather}
      r_{i\,\text{center}} < r_{j\,\text{center}} \\
      [r_{i\min}, r_{i\max}] \nsubseteq [r_{j\min}, r_{j\max}] \\
      [r_{j\min}, r_{j\max}] \nsubseteq [r_{i\min}, r_{i\max}] \\
      [\theta_{i\min}, \theta_{i\min}] \cap [\theta_{j\min}, \theta_{j\max}] \neq \emptyset.
  \end{gather}
  
  Intuitively, $o_i$ is closer to the viewpoint than $o_j$, has overlapping angular bounds, and is not completely within the same radial bounds.
  Equivalently, $o_i$ is left of $o_j$ if
  
  \begin{gather}
      \theta_{i\,\text{center}} > \theta_{j\,\text{center}} \\
      [\theta_{i\min}, \theta_{i\max}] \nsubseteq [\theta_{j\min}, \theta_{j\max}] \\
      [\theta_{j\min}, \theta_{j\max}] \nsubseteq [\theta_{i\min}, \theta_{i\max}] \\
      [r_{i\min}, r_{i\min}] \cap [r_{j\min}, r_{j\max}] \neq \emptyset.
  \end{gather}
  
  Since angles are cyclical, the angular comparison is computed based on the shorter direction of displacement. Although this definition is fairly strict, it reduces the number of extraneous directional relationships.
  
  If viewpoint-dependent relationships are not selected as a constraint, the viewpoint is resampled in subsequent iterations. Otherwise, the viewpoint is not resampled for that query.

  \mypara{Ordinal relationships.}
  Ordinal relationships (\eg first, third) are added for objects $O_c$ of the same class that demonstrate a clear ordering in the scene relative to another \emph{reference} object $o_r$. 
  Distance-based orderings are computed by ranking the distances from each $o\in O_c$ to $o_r$. If the differences between consecutive distances are all above a threshold (\ie no equidistant objects), each $o\in O_c$ is assigned \texttt{n}th closest (or farthest) based on their distance.
  Direction-based orderings are determined relative to a viewpoint. If the angular differences between consecutive $o\in O_c$ relative to $o_r$ are all above a threshold (\ie no collinear objects relative to $o_r$), each $o\in O_c$ is assigned \texttt{n}th from the left (or right) based on their angular ordering. The first $o$ is determined by finding the largest angle gap between consecutive objects.
  To ensure that the ordering is clear, a minimum spacing between objects is required to construct a relative ordering for a group of objects.

  \mypara{Comparison relationships.}
  Comparison relationships include relative distance and height, with comparative (\eg taller) and superlative (\eg tallest) descriptors. These are computed by comparing the distances between bounding boxes of objects to each other reference object in the room. As with ordinal relationships, a threshold enforces a minimum difference in distance or height to ensure that the comparison is visually noticeable.

  \subsection{Prior Work}
  \label{appendix:sg-prior}

    \begin{table*}[tb]
\centering
\resizebox{\linewidth}{!}{
\begin{tabular}{lllllccccc} 
\toprule
& \multicolumn{2}{c}{\textit{Attr}} & \multicolumn{7}{c}{\textit{Relationships}} \\ 
\cmidrule(){2-3} \cmidrule(l){4-10}
Method                                                & approach  & vocab  & approach  & view        & vert         & arr          & ord          & comp         & room          \\ 
\midrule
VL-SAT~\citeyearpar{wang2023vlsat}                    & \myxmark  & -      & learned   & scene       & \mycheckmark & \myxmark     & \myxmark     & \mycheckmark & \myxmark      \\
SceneGraphFusion~\citeyearpar{wu2021scenegraphfusion} & \myxmark  & -      & learned   & scene       & \mycheckmark & \myxmark     & \myxmark     & \mycheckmark & \myxmark      \\ 
Open3DSG~\citeyearpar{koch2024open3dsg}               & embedding & open   & embedding & scene       & \mycheckmark & \myxmark     & \myxmark     & \mycheckmark & \myxmark      \\
HOV-SG~\citeyearpar{werby23hovsg}                     & embedding & open   & geo       & \myxmark    & \myxmark     & \myxmark     & \myxmark     & \myxmark     & \mycheckmark  \\
ConceptGraphs~\citeyearpar{gu2024conceptgraphs}       & VLM       & open   & VLM       & \myxmark    & \mycheckmark & \myxmark     & \myxmark     & \myxmark     & \myxmark      \\
OpenFunGraph~\citeyearpar{zhang2025openvocabulary}    & VLM       & open   & VLM       & \myxmark    & \myxmark     & \myxmark     & \myxmark     & \myxmark     & \myxmark      \\ 
\midrule
3DSSG (SGPN)~\citeyearpar{wald2020learning}           & learned   & closed & learned   & scene       & \mycheckmark & \myxmark     & \myxmark     & \mycheckmark & \myxmark      \\
EmbodiedScan~\citeyearpar{wang2024embodiedscan}       & \myxmark  & -      & geo       & allocentric & \mycheckmark & \mycheckmark & \myxmark     & $\sim$       & \myxmark      \\
SceneVerse~\citeyearpar{jia2024sceneverse}            & VLM       & open   & geo       & scene       & \mycheckmark & \mycheckmark & \myxmark     & \myxmark     & \myxmark      \\
3D-GRAND~\citeyearpar{yang20243d}                     & VLM       & open   & VLM       & scene       & \mycheckmark & \mycheckmark & \myxmark     & \mycheckmark & \mycheckmark  \\ 
\rowcolor{gray!20} \oursshort                         & VLM       & open   & geo, VLM  & agent       & \mycheckmark & \mycheckmark & \mycheckmark & \mycheckmark & \mycheckmark  \\
\bottomrule
\end{tabular}
}
\caption{\textbf{Comparison of scene graph extraction methods.} 
We compare implementations and support for different attribute and relationship types.
The first section shows general methods for scene graph extraction, while the second shows methods used to generate 3DVG datasets.
Different approaches for extracting attributes and relationships include \textbf{geo}metric methods, \textbf{learned} methods to map to a closed set, methods which generate \textbf{embedding} representations, and training-free \textbf{VLM}s.
Attributes can also be closed- or open-vocabulary.
We additionally show which relationship types are supported by each method (\textbf{vert}ical, \textbf{arr}angements, \textbf{ord}inal, \textbf{comp}arisons, and relationships relative to \textbf{room}s).
}
\label{tab:sg-type-comparison}
\end{table*}

    The creation of a 3D scene graph is a common step for automated pipelines that generate data for 3D vision and language tasks.  
    Some works rely on scene graphs provided by the 3D scene dataset~\citep{huang2023embodied,zhu20233d}, while others use 3D geometric heuristics~\citep{achlioptas2020referit3d,etesam20223dvqa,jia2024sceneverse} to obtain relationships between ground-truth objects.
    It is also possible to obtain object relationships after projecting into 2D, either via heuristics~\citep{johnson2017clevr} or by prompting a VLM with the rendered image~\citep{wang2024embodiedscan,yang20243d}. 
    Object attributes are typically obtained from a combination of ground-truth annotations (e.g. semantic class of the object), 3D reasoning (e.g. using the size of the bounding box to estimate the size of the object), and appearance attributes (e.g. color, style) from a 2D vision model.
    In the following, we briefly review prior work on 3D scene graphs and methods for extracting them.
    
    \mypara{3D scene graphs.} A 3D scene graph represents a 3D scene as a structured graph, where nodes correspond to objects and edges capture their relationships, thus facilitating higher-level semantic reasoning about how objects relate to each other.
    In 3D, such scene graphs has been used for representing constraints for 3D scenes~\citep{chang2014learning} and as a sparse semantic representation of a 3D environments~\citep{kim20193}. 
    \citet{armeni2019threedscenegraph} introduced the concept of 3D scene graph as a unified representation across multi-level entities and structures in a 3D building, encompasing objects, rooms, and cameras and their relationships. 
    In our work, we focus on object-centric 3D scene graphs that considers objects and their relationships with other objects and the overall environment (\eg the room).
    
    \mypara{Scene graph extraction.} To extract scene graphs for a 3D scene,
    it is necessary to identify the objects, their attributes, and relationships between the objects.
    Methods for extracting scene graphs typically focus on identifying relationship between objects, and rely on 3D object detectors for identifying objects.  
    Early work in this area used geometric heuristics to extract relationships between objects~\citep{wang2019planit,luo2020end}.  
    \citet{wald2020learning} was one of the first works to train an end-to-end model to predict a semantic 3D scene graph from a point cloud.
    To do so, they contributed 3DSSG, a dataset of 3D scenes with geometrically computed and manually verified 3D scene graphs. 
    Follow-up works show that it is possible to generated scene graphs in real-time from an RGB-D video~\citep{wu2021scenegraphfusion}, and improve scene graph predictions for rare relationships by fusing 2D visual features, language, and 3D geometry~\citep{wang2023vlsat}.
    These methods that relies on training supervised models based on labeled data are restricted to attributes and relationships that are defined in the dataset.
    More recent works~\citep{gu2024conceptgraphs,koch2024open3dsg,werby23hovsg,zhang2025openvocabulary} leveraged advances in object detectors and large-scale VLMs to obtain scene-graphs with open-vocabulary labels for objects and relationships. 
    Despite these advances, existing 3D scene graph generation methods often have narrow focuses in terms of attribute or relationship types. For instance, 3DSSG~\citep{wald2020learning} is limited to a closed set of attributes and relationships. ConceptGraphs~\citep{gu2024conceptgraphs} uses a non-exhaustive captioning approach for attributes and focuses exclusively on supporting relationships (\emph{in} and \emph{on}), and HOV-SG~\citep{werby23hovsg} uses CLIP embeddings to represent attributes, which are not good at aligning with fine-grained text descriptions, and only consider room assignment relationships.
    In contrast, our approach enriches the scene graph by embedding consistent object attributes alongside semantic relationships, enabling more detailed and actionable scene understanding.
    We compare our implementation with other methods, including the attribute and relationship types that each supports, in \cref{tab:sg-type-comparison}.

    \mypara{Scene graph approaches for 3DVG datasets.} 3D-VisTA~\citep{zhu20233d} uses ground truth annotations from 3DSSG, while LEO-Align~\citep{huang2023embodied} extracts scene graphs using the trained 3DSSG model. These are limited since 3DSSG is trained only on 3RScan, and the ground truth set of attributes and relationships are limited in diversity and not error-free. 
    Several works~\citep{jia2024sceneverse,yang20243d} use VLMs to extract attributes, either by list or caption. 
    To extract object relationships, 3D-GRAND uses a VLM, while EmbodiedScan~\citep{wang2024embodiedscan} and SceneVerse uses geometric rules to predict spatial relationships. 
    We find that these also are limited in precision and diversity. For instance, the methods do not cross-reference attributes across objects, and various relationships are defined in restricted forms (\eg SceneVerse generates aligned relationships but only along the $x$ and $y$ axes). We improve on these VLM and geometry-based methods by using cross-referencing to ensure consistency across attribute assignment, expanding the supported attribute and relationship types, and implementing our method to support scalability across scene datasets and types.

  \subsection{Evaluation}
    \label{appendix:sg-evaluation}
    \begin{table}[t]
\parbox{.45\linewidth}{
    \centering
    \begin{tabular}{llr}
    \toprule
    \textbf{Metric} & \textbf{Value} \\
    \midrule
    Precision & $80.5$ \\
    Recall & $57.0$ \\
    F1 Score & $63.3$ \\
    \bottomrule
    \end{tabular}
    \caption{\textbf{Similar object prediction}. Our method predicts similar objects conservatively to achieve high precision while reducing the chance of non-similar objects being assigned the exact same attributes.}
    \label{tab:scenegraph_results}
}
\hfill
\parbox{.45\linewidth}{
    \centering
    \begin{tabular}{lrrr}
    \toprule
    \textbf{Method} & \textbf{CLIP} $\uparrow$ & \textbf{Dice} $\uparrow$ & \textbf{IoU} $\uparrow$ \\
    \midrule
    GPT-4.1 & 0.902 & 0.699 & 0.579 \\
    \rowcolor{gray!20} \oursshort & \best{0.943} & \best{0.910} & \best{0.864} \\
    \bottomrule
    \end{tabular}
    \caption{\textbf{Attribute consistency} for 60 human-annotated similar object pairs. The predicted attributes from \oursshort between similar objects are more consistent compared with a baseline approach that directly predicts attributes per object.}
    \label{tab:similarity-consistency}
}
\end{table}

    To evaluate the quality of our extracted scene graphs, we conducted a user study where we ask participants to validate each predicted attribute and relationship for 14 scenes. 
    For each scene, we sampled 15 objects and considered only their attributes and relationships among the sampled objects, since it was infeasible to manually evaluate the full scene graphs. 
    External annotators validated the predicted attributes and relationships of each object and also annotated similar object pairs. 
    Results are shown in~\cref{tab:scenegraph_results}.

    \mypara{Attribute consistency.} We first evaluate whether \emph{cross-referencing} improves the quality of our scene graphs, by evaluating 1) how well our method identifies similar objects and 2) whether identical attributes are assigned to similar objects.
    Across 14 scenes and subsampled objects, users identified 60 similar object pairs in total, including 41 from ScanNet scenes and 19 from 3RScan scenes.
    
    For 1), we compare the predicted similar-object pairs against human-annotated pairs and report precision, recall, and F1, restricting evaluation to pairs involving the sampled objects. The high precision (80.5) but moderate recall (57.0) indicates conservative similarity predictions that favor correctness over coverage.
    
    For 2), we share the results in \cref{tab:similarity-consistency}, measuring consistency through the CLIP score between attribute assignments of similar objects, the Dice coefficient (2 times the number of shared attributes over total non-unique attributes), and the IoU (number of shared attributes over total unique attributes). We compare against a baseline method that directly predicts attributes with a similar prompt to \cref{prompt:attribute-extraction}. We find that our method achieves significantly higher consistency of attributes, suggesting that cross-referencing helps our method extract scene graphs with greater attribute consistency.

    \mypara{Attribute evaluation.}
    For attributes, we compute the CLIP~\citep{radford2021learning} image-text cosine similarity between an object-centric RGB crop and the predicted attributes. We additionally compute the CLIP Score\textsuperscript{*} using the human-validated attribute set (\ie predicted attributes marked correct by annotators), which serves as an upper bound. The predicted attributes achieve a CLIP score of 0.245 comapred to the human-validated score of 0.249, suggesting accurate attribute descriptions.
    
    \mypara{Relationship evaluation.}
    We find through our user study that our extracted relationships have a precision of 89.7\%, suggesting that our method has reliable relationship prediction.
    
    \mypara{Comparison against scene graph extraction methods.}
    We omit direct comparisons against prior methods for several reasons.
    For one, existing methods do not support as many attribute and relationship types as \oursshort, so performance on several types cannot be compared.
    Second, it is infeasible to produce a comprehensive ground truth of attributes and relationships, due to the plethora of attributes and relationships and differences in vocabulary (\eg turquoise vs. light blue) and variability in definitions (\eg what constitutes \emph{far}) that make the evaluation of open-vocabulary methods challenging or expensive. 
    Finally, many of the learned methods~\citep{wald2020learning,wang2023vlsat,zemskova20253dgraphllm} are trained and evaluated on 3DSSG~\citep{Wald2019RIO}, which we found to have errors and a limited view of the attributes and relationships. 
    Thus, it is not appropriate to compare our method or other open-vocabulary methods to these learned methods.

\clearpage
\section{Constraint Solver}
  \label{appendix:constraint-solver}
  \begin{listing*}[t]
  \centering
  \begin{lstlisting}[language=Python,frame=single,breaklines=true]
"""
### INSTRUCTIONS ###
You are an agent in a scene trying to come up with a way to describe what object to find and where it is, using the given constraint descriptions. You should use *all* of the provided constraints to craft a combined visual grounding description, and you should not add any additional details.
The description must be a natural rephrasing of the provided template sentences, maintaining an equivalent level of specificity. 

Requirements:
- You must use *every* constraint (barring duplicates) in the description and refer at least once to *every* mentioned object in the constraints. Since there can be more than one of each object in the scene, any constraints describing objects other than the target are also necessary to disambiguate them from others of the same class in the scene.
- When referring to objects, you should use articles ("a", "an", "the", "another") as needed to make it clear which objects you are referring to. If multiple objects of the same class are mentioned, be sure to clarify which is which using the constraints. Specifically, if the constraints describe anchor objects of the same class as the target, make it clear which of the objects of that class is the target. Examples: if the constraint is "target = <chair:15,16>", then your sentence can say "The target is a <chair:15,16>."
- The target object should be clear from the description. For instance, if there is a target object A next to object B, then you could say (but are not limited to) one of the following:
  - "A is next to B."
  - "A can be found next to B."
  - "This is A next to B..."
  - "Next to B is A..."
  - "Look for A..."
  - "...the object of interest A..."
  - "Look next to B. The object you are looking for is A."
- When referring to the same object multiple times, you can use pronouns or synonyms to avoid repetition, but it must still be clear which object you are referring to.
- If 'you' is included in the constraints, describe the viewpoint from which the relationships are being grounded. The 'you' can be implied or explicit, e.g. "when standing in front of the couch", "from the corner of the room", "when you are at the center of the room", "to your left", or "behind you".
- If "not" is used in a constraint (e.g. "<thing:24> -> not in -> <corner:4542>"), you must also use the word "not" in the generated description (e.g. "the thing of interest is not in the corner").
- Do not infer any additional relationships or attributes that are not explicitly provided in the constraints. For instance, if object A is described as between B and C, do not assume that A is near B or C unless specified.
- Do not explicitly reference the type of the attribute in the final description UNLESS it is necessary to clarify what the attribute is referring to. For instance, adjectives such as "standard" and "light" are ambiguous and should be clarified naturally (e.g. "standard-sized", "light-colored").
- The candidate descriptions should include the ID of each mentioned object to distinguish different referents. Use the syntax "<object name:X>", where X is the integer ID number. If a noun refers to multiple objects, specify this with commas as <object name:X,Y,Z>. In your final output description, please use the corresponding ID tags for all references to objects, including coreferences.

### EXAMPLE FINAL DESCRIPTIONS ###
Use these descriptions as examples for the diversity of acceptable language:
- <This:3> is a cardboard <box:3> against the wall and right next to two <doors:11,13>. Of such <boxes:3,4,5>, it is the <one:3> not next to an <armchair:28>.
- On the brown, wooden <dining table:6> is a small, rectangular <projector:7>.
- <This:5> is the larger of the two <toolboxes:5,6> near the <piano:10>.
- Find the <object:11> with wheels.
- When facing the <radiator:41>, the <metal rail:37> is directly on your left.
- If you stand in front of the <radiator:42>, the <rail:1> is right above <it:42>. <It:1> should be made of metal.
- To the left of the <sink:4> is a set of <paper towel rolls:101,102>.
- This circular <object:18,23> is supported by a white <mini fridge:17>.
- Underneath the <sink:23> is a <cabinet:31> for storing cleaning supplies.
- Locate the tallest <appliance:25> in the kitchen.
- There is a row of black rolling <chairs:3,4,5,6> along the <wall:29>. Pick the <one:5> furthest to the left when facing <them:3,4,5,6>.

Each description has its own scene and distinct IDs, so do NOT use any of the relationships or IDs based on the examples.
"""
  \end{lstlisting}
  \caption{\textbf{Rephrasing Prompt.} To generate a first iteration of the grounding description, we pass the target name, sampled constraints, and optional viewpoint description to a VLM and prompt it to generate a naturally phrased description encompassing all of the constraints.
  }
  \label{prompt:rephrase}
\end{listing*}

  \begin{listing*}[t]
  \centering
  \begin{lstlisting}[language=Python,frame=single,breaklines=true]
"""
### INSTRUCTIONS ###
You are a helpful assistant well versed in English grammar. Your task is to correct the grammar of the provided sentence and make it more natural while preserving their original meaning.

Requirements:
- Do not change the semantic meaning or add any new information about the objects. It is preferred to keep the word choice close to the original but rearrange the words, change articles or prepositions, or choose better synonyms to make it more naturally sounding. 
- If a sentence is a run-on or has many clauses, split it into multiple sentences. 
- If an object is repeated many times, you can also substitute with pronouns (preserve the ID tags).
- If the sentence is fine as is, simply return it unchanged. You should return only the corrected sentence without any additional text or explanations.

Keep any ID tags such as "<object name:X,Y>" as-is, and do not change any of the brackets. Use the ID tags as clues for which nouns or pronouns refer to the same object.

Grammar corrections should include the following:
* Add or correct any articles ("the", "a", "an", "another") to refer to each noun. If the same object name appears multiple times but applying to different object IDs, use articles and conjunctions to specify that they are explicitly different (e.g. "a", "another", "a different", "one of"). For instance, if the sentence talks about "the <thing:X>" and later "the <thing:Y>" with a different ID, then you could refer to the second <thing:Y> as "a different <thing:Y>".
* If any of the word choices are awkward or unnatural, replace them with more natural synonyms (e.g. "an object used for sitting" can be changed to "an object that someone can sit on"). Otherwise, try to keep the original words as much as possible.
* If there is a number between 0 and 9 in a prompt (e.g. 3), change it to the corresponding word ("three") and make sure that the pluralization is correct (e.g. "5 toy" should be "five toys").
"""
  \end{lstlisting}
  \caption{\textbf{Grammar correction prompt.} We pass the generated visual grounding description back to the LLM to improve readability, including treatment of articles, word choice, and sentence structure, while preserving the underlying semantics of the description.
  }
  \label{prompt:grammar-correction}
\end{listing*}

  The prompts for rephrasing and grammar correction are provided in \cref{prompt:rephrase} and \cref{prompt:grammar-correction}, respectively. We provide further details regarding the implementation of the constraint solver below.

  \subsection{Constraint Sampling}

  Rather than a naive uniform sampling of constraints, we modify sampling in three ways. First, we prioritize sampling of constraints to those that reduce the pool of distractors by at least half. Second, to keep the solution space from exploding in size when relationships (and thus new referents) are added, in the last step of sampling, we set the unnormalized probability of each new relationship constraint based on the number of objects with the constraint predicate label $n_o$ and the current number of sampled constraints $n_c$:
  \begin{equation}\label{eq:constraintWeighting}
      P(n_o, n_c) \sim \sigma\left(-\frac{n_c}{k} \left( n_o - \left(n_0 + \frac{n_1 - n_0}{2\sqrt{n_c}}\right) \right)\right),
  \end{equation}
  where $\sigma(x)$ is the sigmoid function and $k$ is a scaling factor given by
  \begin{equation}
      k = \frac{n_1 - n_0}{2(\ln(1-\tau) - \ln\tau)}.
  \end{equation}
  Here $\tau$ is a value between 0 and 1 such that the probability with $n_c=1$ of sampling a constraint with $n_o = n_1$ is $\tau$, and the probability of sampling a constraint with $n_o = n_0$ is $1 - \tau$. If $n_c = 0$, then we assign an equal weight to all constraints. Intuitively, the more constraints that have already been sampled, the more sampling will preference constraints that minimally introduce new avenues of exploration, to prevent the solution from exploding in size.

  \subsection{Terminal Condition}
  \label{appendix:terminal-condition}

  To terminate constraint solving for multi-target queries, we use a similar formulation to \cref{eq:constraintWeighting} based on the current number of targets $n_t$:
  \begin{equation}
      P(n_t) \sim \sigma\left(\frac{1}{\kappa} \left( n_t - \left(n_{t0} + \frac{n_{t1} - n_{t0}}{2}\right) \right)\right),
  \end{equation}
  where $n_{t0}$, $n_{t1}$, and $\kappa$ are defined analogously. This intuitively limits the number of targets in multi-target queries to reasonable numbers.

\section{Dataset Statistics and Quality}
  \label{appendix:dataset}

  \begin{table*}[htbp]
\centering
\resizebox{\linewidth}{!}
{
\begin{tabular}{@{}llr@{}}
\toprule
Variable & Description & Value \\
\midrule

$p_0$ & probability of zero-target query & 0.2 \\
$p_1$ & probability of single-target query & 0.5 \\
$p_{>1}$ & probability of multi-target query & 0.3 \\
$p_{\text{lab}}$ & probability of sampling unique label & 0.1 \\
$p_{\text{attr}}$ & probability to sample an attribute (vs. relationship) & 0.5 \\
$p_{\text{neg}}$ & probability to sample negative constraint & 0.1 \\
$n_{\text{constraints}}$ & maximum number of constraints in description & 8 \\
$n_{\text{anchors}}$ & maximum number of anchors before guaranteed exploration & 3 \\
$n_0$ & number of objects at which constraint is sampled with $1-\tau$ probability ($n_c=1$) & 3 \\
$n_1$ & number of objects at which constraint is sampled with $\tau$ probability ($n_c=1$) & 6 \\
$\tau$ & probability for constraint sampling & 0.05 \\
$n_{t0}$ & number of targets at which multi-target sampling terminates with probability $\tau_{\text{term}}$ & 2 \\
$n_{t1}$ & number of targets at which multi-target sampling terminates with probability $1-\tau_{\text{term}}$ & 5 \\
$\tau_{\text{term}}$ & probability for multi-target sampling termination & 0.95 \\
\midrule

$s_{\text{layout}}$ & grid resolution for scene layout & 0.05 \\
$n_{\text{room}}$ & threshold for min number of grid cells for room & 100 \\
$r_{\text{layout}}$ & dilation radius for wall cells & 2 \\
\midrule

$n_{\text{views}}$ & number of views given to VLM for attribute extraction & 3 \\
$A_{\text{thresh}}$ & area percentile threshold for image sampling & 75 \\
$h_{\text{image}}$ & maximum dimension for image & 384 \\
$m_{\text{crop}}$ & relative margin for cropping of object & 0.3 \\
\midrule

$d_{\text{next}}$ & maximum distance for next relationship & 0.4572 \\
$d_{\text{near}}$ & maximum distance for near relationship & 1.2192 \\
$t_{\text{near}}$ & relative threshold for near relationship & 0.25 \\
$d_{\text{far}}$ & minimum distance for far relationship & 2.1336 \\
$d_{\text{corner}}$ & threshold for in-corner relationship & 0.5 \\
$d_{\text{center}}$ & threshold for in-center relationship & 0.8 \\
$d_{\text{betweeen}}$ & distance between objects to be considered ``in between'' & 0.1 \\
$\theta_{\text{between}}$ & max angle displacement among three objects to be considered ``in between'' & 0.1745 \\
$d_{\text{align}}$ & distance between objects to be considered ``aligned'' & 0.2 \\
$d_{\text{ordinal}}$ & distance between objects to be considered for ordinal relationships & 0.25 \\
$t_{\text{comparison}}$ & relative threshold for comparison relationship & 0.25 \\ 
\bottomrule
\end{tabular}
}
\caption{\textbf{Hyperparameters of \oursshort.} We use the following settings to generate the \oursshort dataset. All distances are in meters, and all angles are in radians.}
\label{tab:dataset-hyperparameters}
\end{table*}

    \begin{figure*}[t]
  \centering
  \includegraphics[trim={0 0 0 4px},clip,width=\linewidth]{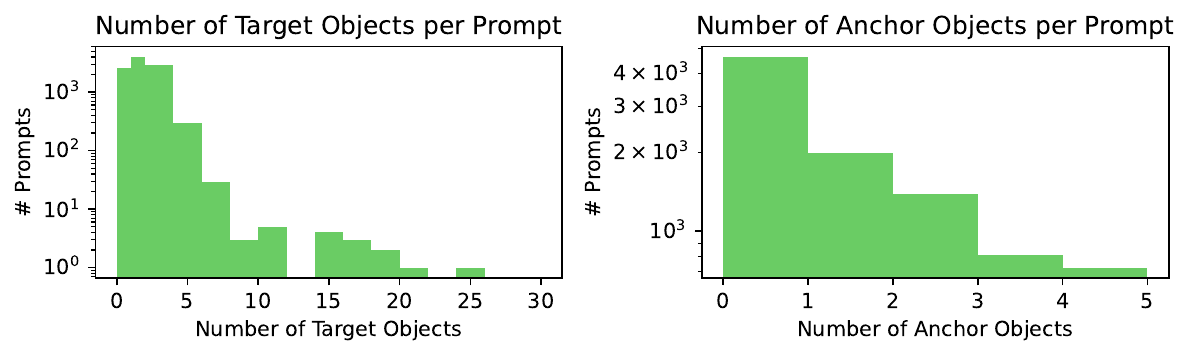}
  \caption{\textbf{Object counts.} Number of target objects (left) and anchor objects (right) per VG query in \oursshort.
  }
  \label{fig:distribution-objects}
\end{figure*}

    \begin{figure*}[t]
  \centering
  \includegraphics[trim={0 0 0 4px},clip,width=\linewidth]{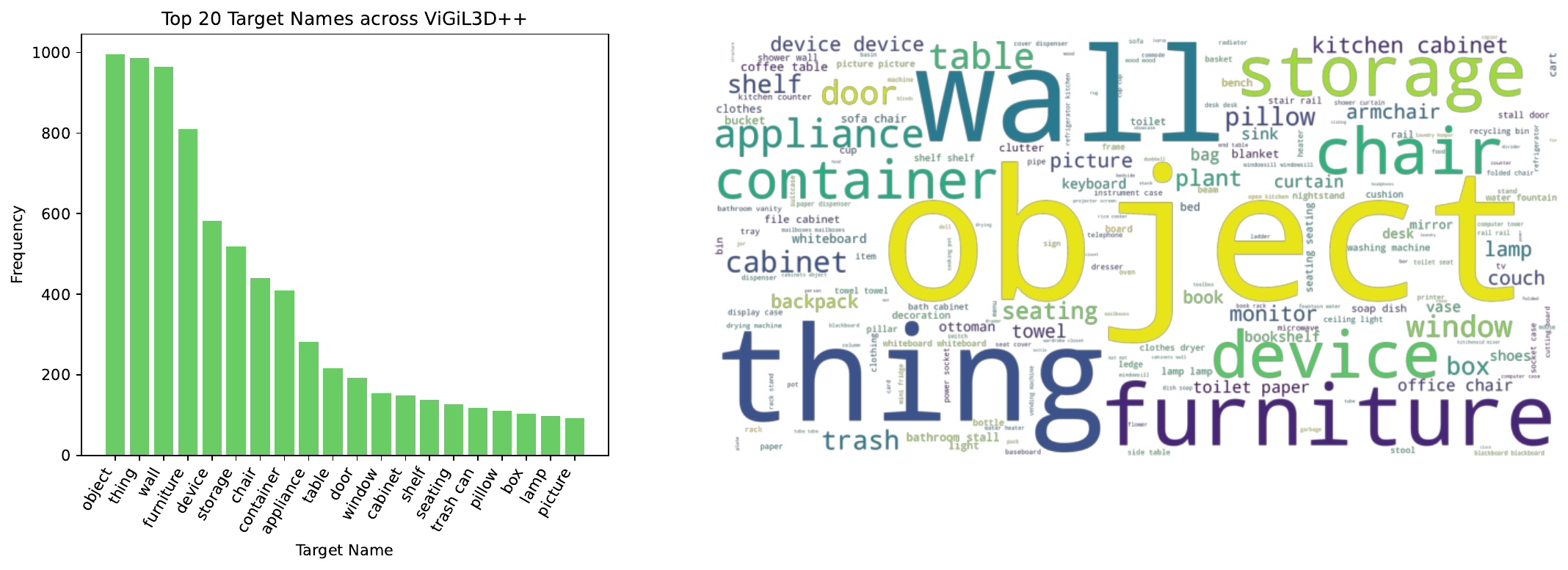}
  \caption{\textbf{Target object word cloud.} Frequency of target object referents in \oursshort. The most frequent target labels are more coarse-grained, making the queries more challenging and requiring models to not only rely on unique object names for grounding.
  }
  \label{fig:target-word-cloud}
\end{figure*}

    \begin{figure*}[t]
  \centering
  \includegraphics[trim={0 0 0 4px},clip,width=\linewidth]{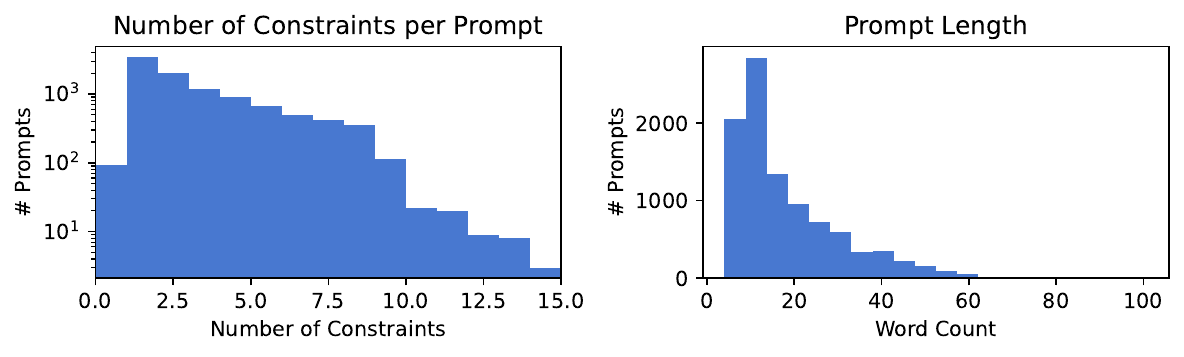}
  \caption{\textbf{Query complexity.} Number of sampled constraints (left) and number of words (right) per VG query in \oursshort.
  }
  \label{fig:distribution-lengths}
\end{figure*}

  \begin{figure}[ht]
  \centering
  \setkeys{Gin}{width=\linewidth}
  \begin{tabularx}{\textwidth}{Y@{\hspace{1mm}} l m{2in} m{1.5in}}
    \toprule
    GT & Method & Description & Failure \\
    \midrule
    \includegraphics{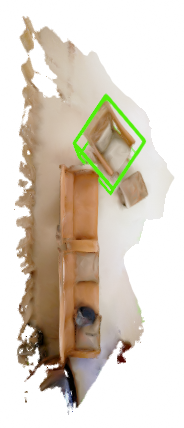} & Im2Query & Find \emph{the piece of furniture} with wide wooden arms and tan fabric cushions covered in leafy motifs, positioned next to the floor tiles. & The large couch with pillows on it can also be described by this in the scene. \\
    \includegraphics{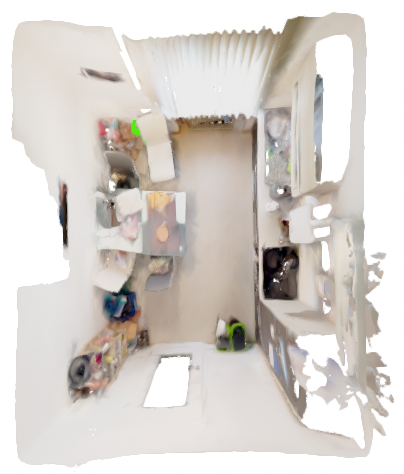} & Cap2Query & These are the \emph{two} tall, thin boxes with colorful printed designs, standing upright and leaning in the corner where the white wall and wooden baseboard meet, directly behind the white chair \emph{with blue trim} and beside a metal chair leg on the beige carpet. & There is only one box in the scene, and the chair does not have blue trim. \\
    \includegraphics{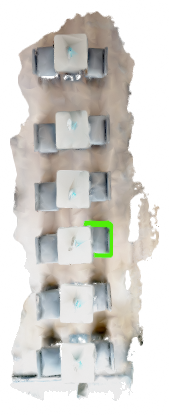} & SG2Query & This is a black, rectangular, leather chair located in the center of the restaurant, next to both the table and table and in between table and table. It is the only chair that is next to \emph{table, table, table, chair, chair, and chair} all together. & The chair is not uniquely described, and many of the referents are ambiguous and grammatically incorrect. \\
    \includegraphics{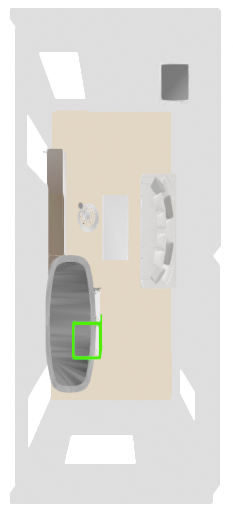} & 3D-GRAND & This is a \emph{dining chair} with a sleek ivory-colored upholstery that exudes sophistication. The chair has an angular, high-back design that offers a contemporary silhouette. Positioned next to the dining table and \emph{behind} the TV stand, the ivory dining chair adds a touch of elegance to the dining area. & The dining chair is not uniquely described, and the chairs are not behind the TV stand. \\
    \bottomrule
  \end{tabularx}
\captionof{figure}{\textbf{Failure cases for prior methods}. Previous methods most often fail to distinguish the target object from distractors.}
\label{fig:failure-cases}
\end{figure}

  \begin{figure}[ht]
  \centering
  \setkeys{Gin}{width=\linewidth}
  \begin{tabularx}{\textwidth}{Y@{\hspace{1mm}} m{2in} m{2in}}
    \toprule
    GT & Description & Failure \\
    \midrule
    \includegraphics{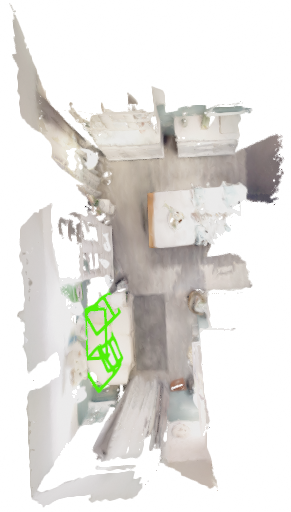} & Look for a \emph{light green} pillow that is far from a corner. & Not all of the pillows are light green. Attributes can be difficult to parse due to VLM errors and challenging lighting conditions. \\
    \includegraphics{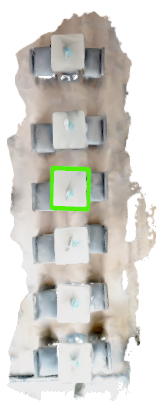} & From the center of the restaurant, look for the smooth, black table that is used for eating and for placing objects on. This table is \emph{higher than the window} and is not in front of it. & Due to an imperfect reconstruction, the table appears higher than the window. This error is repeated across several descriptions, as the scene is homogeneous (same chairs and tables and few distinct landmarks in the scene). \\
    \includegraphics{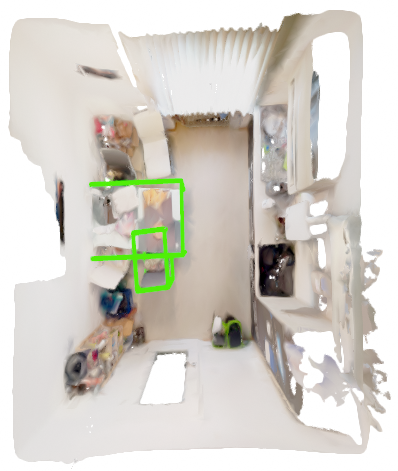} & Look for the furniture in the kitchen that is next to a lamp and \emph{near a trash can}. & ``Near'' is hard to judge, since the room is small, and there is a noticeable gap between the trash can and furniture. In other contexts, this distance may be considered within the threshold. \\
    \includegraphics{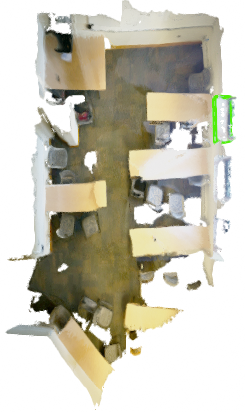} & When you are in the center of the office, the windowsill supporting a window is \emph{to the right of a whiteboard}. & There are multiple windowsills that are technically to the right of the whiteboard. However, it can be up to interpretation as to how long-distance directional relationships can apply. \\
    \bottomrule
  \end{tabularx}
\captionof{figure}{\textbf{Failure cases for \oursshort}. Common validity failures include incorrect attributes or relationships, imperfect point cloud reconstructions, and contextual relationships.}
\label{fig:failure-cases-vigil3dpp}
\end{figure}

  \begin{table*}[tb]
\centering
\resizebox{\linewidth}{!}
{
\begin{tabular}{@{}l rrrrrr @{}}
\toprule
Model & Avg. Length & CR ($\downarrow$) & NGD ($\uparrow$) & Self-Rep. ($\downarrow$) & Hom. (R-L) ($\downarrow$)) & Hom. (BERT) ($\downarrow$) \\
\midrule
\oursshortman~\citeyearpar{wang2024vigil3d} & 14.2 & 3.387 & 2.340 & 1.197 & 0.187 & 0.615 \\
\midrule
3D-GRAND~\citeyearpar{yang20243d} & 44.6 & 9.406 & 0.637 & 6.117 & 0.312 & 0.698 \\ %
ScanScribe~\citeyearpar{zhu20233d} & 35.7 & 6.339 & 1.087 & 6.051 & 0.266 & \best{0.636} \\
SceneVerse~\citeyearpar{jia2024sceneverse} & 9.8 & \best{5.011} & \best{1.512} & \best{2.812} & 0.282 & 0.657 \\ %
\rowcolor{gray!20} \oursshort & 18.6 & \second{5.632} & \second{1.207} & \second{4.048} & \best{0.265} & \second{0.638} \\
\bottomrule
\end{tabular}
}
\caption{\textbf{Dataset diversity}. We compare the linguistic diversity of \oursshort to prior LLM-scaled datasets using the \texttt{diversity}~\citep{shaib2024standardizing} package, which measures the amount of redundancy in a linguistic text. Metrics include the Compression Ratio (CR) of the compressed (gzip) over the original size, the $N$-Gram Diversity Score (NGD), the Self-Repetition Score (Self-Rep), and Homogenization Score (using Rouge-L and BERTScore) to measure pairwise document similarity. While these metrics can be biased by the text length, we find that \oursshort beats most MLLM-scaled baselines.
}
\label{tab:dataset-diversity}
\vspace{-3mm}
\end{table*}

    \begin{figure*}[t]
  \centering
  \includegraphics[trim={0 0 0 4px},clip,width=\linewidth]{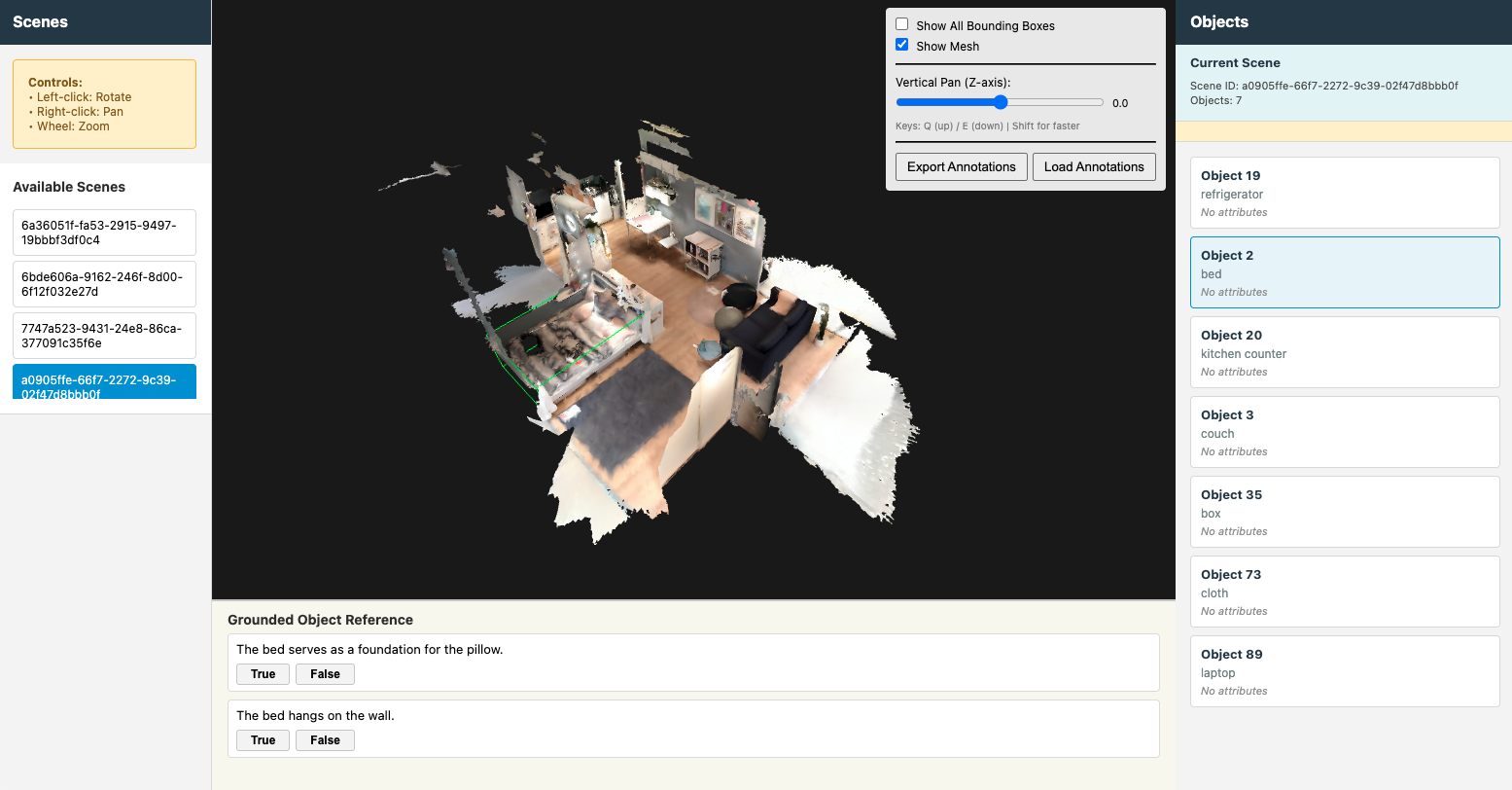}
  \caption{\textbf{User study interface.} Users are presented with the 3D point cloud of the scene (center) and a grounding description (bottom) along with the localization of the target(s). The proposed targets are highlighted in the bounding box (\textcolor{green}{green}). If the targets are exactly the objects described by the query, then the user selects ``True''. Otherwise, the user selects ``False''.
  }
  \label{fig:validation-interface}
\end{figure*}

  \begin{table}[tb]
\centering
\resizebox{\linewidth}{!}
{
\begin{tabular}{@{}lrrrrrr@{}}
\toprule
& Zero & \multicolumn{3}{c}{Single} & Multi & Total \\
\cmidrule(){2-2} \cmidrule(l){3-5} \cmidrule(l){6-6} \cmidrule(l){7-7} 
VLM & & Unique & Common & Total & & \\
\midrule
GPT-4.1~\citeyearpar{achiam2023gpt} & \best{78.6 (9.6)} & \best{100.0 (0.0)} & \best{43.1 (12.0)} & 47.1 (11.7) & \best{45.7 (11.7)} & \best{57.1 (6.7)} \\
InternVL~\citeyearpar{wang2025internvl3_5} & 50.0 (13.9) & 75.0(30.0) & 23.8(12.9) & 32.0 (12.9) & 30.0 (12.7) & 37.3(7.7) \\
Qwen3-VL~\citeyearpar{qwen3technicalreport} & 72.0 (12.4) & 83.3(21.1) & 36.8(15.3) & \best{48.0 (13.8)} & 36.0 (13.3) & 52.0(8.0) \\
\bottomrule
\end{tabular}
}
\caption{\textbf{Query validation with different VLMs.} We evaluate the correctness of generated queries for \oursshort using different VLMs, including both proprietary and open-weight models. For zero-target queries, we measure the proportion of queries that correctly do not describe any object in the scene. For single- or multi-target queries, we measure correctness as the proportion of queries that describe only the proposed targets. We split correctness by targets that are \emph{unique} (Uni) in their object class vs. \emph{common} (Com). We observe decreased performance with both InternVL and Qwen3-VL in most metrics.}
\label{tab:prompt-validation-vlms}
\end{table}

  \lstdefinestyle{plaintext}{
  basicstyle=\tiny\ttfamily,
  showstringspaces=false,
  breaklines=true,
  language=,
  numbers=none,
  keywordstyle=,
  commentstyle=,
  stringstyle=,
}

\begin{listing*}[t]
  \centering
  \begin{lstlisting}[frame=single,breaklines=true,style=plaintext]
  Given a 3D representation of a scene (e.g. a room) and a description of an entity (or entities) within the scene, visual grounding is the problem of localizing the target object(s) being referred to by the description. For instance, given the example below and a description such as, "This is the pillow on the couch that is closest to the black bag," the model should identify the pillow annotated below.

  For each presented prompt, your goal is to validate whether the visual grounding prompt is correctly describing the shown target object(s) (if any). Some high-level details:
  * We are only considering objects which are pre-segmented and presented for you in the interface. For the purposes of validation, assume that the segmentation labels are correct
  * Each visual grounding prompt in the ViGiL3D dataset can be a description of zero, one, or multiple objects in the provided scene (examples shown below). 
    * Zero objects => no object in the scene matches the description (nothing should be highlighted, and the list of object IDs should be empty)
    * One object => exactly one object in the scene matches the description
    * Multiple objects => multiple objects satisfy the description
  * You will be presented with the ground truth objects. Please verify whether 1) those objects are correctly described by the prompt and 2) whether those are the only objects described by the prompt.
  * You should only need the rendered 3D point cloud in the browser to identify the appropriate targets for each object, but you may still find it useful as well to view the corresponding RGB-D frames.


  Directions:
  1. Go to <URL> to view the scene assignments.
  2. In the scene viewer, select the corresponding scene ID from the left bar.
  3. Click on an object in the ``Objects'' tab on the right to pull up all of the descriptions of that object in the scene. The object should be marked with a green bounding box in the visualizer.
  4. For multi-target prompts, click on ``Multi-Target Prompts'' on the right side and then click on each prompt to see its corresponding targets.
  5. For zero-target prompts, no objects will be marked in the scene.
  6. For each description, select ``True'' if the prompt correctly describes every annotated target (and no other segmented objects in the scene), or ``False'' otherwise.
  7. Export and upload your annotation at <URL>.
  \end{lstlisting}
  \caption{\textbf{User study instructions.}
  }
  \label{user-study-instructions}
\end{listing*}

  The full hyperparameter settings for the \oursshort dataset can be found in \autoref{tab:dataset-hyperparameters}. We configure the parameters to generate comprehensible yet challenging queries with a balanced representation of different types. In particular, we make these queries challenging through the following.
  1) We sample queries to minimize cases where the target class makes it trivial as to which object is being referred (\ie minimizing $p_{\text{lab}}$).
  2) We minimize extraneous constraints: every constraint in a query must be used by the model to identify the target object.
  3) For zero-target queries, we use attributes and relationships describing objects of the target class, rather than generating descriptions of objects trivially not in the scene.

  We report object counts, including target and anchor objects, in \autoref{fig:distribution-objects}. Most queries have between zero and three targets, with single-target queries being the most frequent. 
  Many of the queries with more targets are of large scenes, referring to the target with generic labels (\eg ``object'') and using a low number of constraints. 
  The number of anchor objects, on the other hand, ranges from zero to five reference objects. Since multi-target queries are generated from constraints that must apply to all objects, it is expected that the number of anchor objects that share the same spatial relationship with all targets will be low. 
  We also show the most frequent target object names in \autoref{fig:target-word-cloud}. The most frequent target class names include common object class names such as ``chair'' and ``table'' as well as more generic labels such as ``object,'' ``furniture,'' and ``thing.'' As the latter are widely applicable labels, they were frequently sampled as good candidates for formulating challenging queries.

  The distribution of the number of constraints and the word count per query are shown in \autoref{fig:distribution-lengths}. Queries from \oursshort were generated using 0 to at most 14 constraints, with the majority having up to 8. Queries with many constraints usually involved a reference viewpoint, since both the target object and the reference viewpoint had to be grounded in the scene. Additionally, queries had an average length of 19 words but ranged from 4 to over 100 in outlier cases.

  \subsection{User Study for Dataset Quality}
  \label{appendix:user-study}

  In the user study, we recruited four participants with previous familiarity with 3D scenes to validate the ground truth of 200 queries each for \oursshort and 3D-GRAND across roughly 30 scenes each. Users were provided with a custom web interface to validate scenes, as shown in \autoref{fig:validation-interface}. For each query, users were provided with the navigable 3D point cloud of the scene and the corresponding instance segmentation and the grounding description with the targets visualized in the scene, with optional access to the RGB video when available. The users were then instructed to mark the query as correct if the annotated targets were precisely the objects described by the query or incorrect otherwise. The queries could consist of either zero, one, or more than one target. The full user study instructions can be found in \autoref{user-study-instructions}.

  \subsection{Dataset Experiments}
  \label{appendix:ablation}

  We run several ablation studies to determine the impact of several design decisions on the overall performance of \oursshort. 
  
  \mypara{Choice of VLM.} In addition to GPT-4.1, we experiment with other open-weight VLMs, namely \texttt{InternVL3\_5-30B-A3B}~\citep{wang2025internvl3_5} and \texttt{Qwen3-VL-30B-A3B-Instruct-FP8}~\citep{qwen3technicalreport}, to investigate the comparative performance of alternative models. We provide the validity results in \cref{tab:prompt-validation-vlms}. While both models prove capable of generating VG queries, we find that both models achieve lower validity rates compared to GPT-4.1, observing more frequent errors in following the prompt instructions. This may be reflective of the relative sizes of the model or the need for further development of open-source models, in addition to the lack of translation of common VLM benchmarks to a task such as visual grounding query generation.

  \mypara{Failure cases.} We show failure cases for each of the baseline methods and \oursshort in \cref{fig:failure-cases}.

  \mypara{Diversity analysis.} We additionally analyze the linguistic diversity of \oursshort using the \texttt{diversity}~\citep{shaib2024standardizing} package, providing an alternative method of assessing the diversity and repetition of each dataset. We show the results in \cref{tab:dataset-diversity}. \oursmodel outperforms most scaled baselines except for SceneVerse, which is favored in these metrics due to its significantly shorter queries.

\section{\oursmodel Model Details}
  \label{appendix:model}
  \begin{table*}[tb]
\centering
\resizebox{\linewidth}{!}
{
\begin{tabular}{@{}ll rrrrrr @{}}
\toprule
&  & \multicolumn{2}{c}{Unique} & \multicolumn{2}{c}{Multiple} & \multicolumn{2}{c}{Overall} \\
& Trained & Acc@25 & Acc@50 & Acc@25 & Acc@50 & Acc@25 & Acc@50 \\
\midrule
3D-VisTA & ScanScribe & 46.1 & 43.6 & 20.3 & 18.4 & 25.3 & 23.3 \\
GPS & SceneVerse & \best{50.1} & \best{46.8} & 19.9 & 17.7 & 25.8 & 23.4 \\
\midrule
\oursmodel & \ourdataset & 44.0 & 41.3 & 17.4 & 15.6 & 22.5 & 20.6 \\
\oursmodel & \ourdatasetlarge & \second{48.0} & \second{45.3} & \best{22.8} & \best{20.8} & \best{27.7} & \best{25.5} \\
\bottomrule
\end{tabular}
}
\caption{Accuracy (\%) on ScanRefer using predicted boxes from Mask3D~\citep{schult2023mask3d}. \oursmodel achieves the comparable or better performance across benchmarks, especially on \texttt{multiple} cases.
}
\label{tab:model-evaluation-all-pred-scanrefer}
\end{table*}

  \begin{table*}[tb]
\centering
\resizebox{\linewidth}{!}
{
\begin{tabular}{@{}ll rr rrrrrr @{}}
\toprule
& & \multicolumn{2}{c}{\oursshortman (ScanNet)} & \multicolumn{6}{c}{F1 on Multi3DRefer} \\
\cmidrule(){3-4} \cmidrule(l){5-10}
& Trained & Acc@50 & F1@50 & ZT w/o D & ZT w/ D & ST w/o D & ST w/ D & MT & All \\
\midrule
3D-VisTA & ScanScribe & 13.4 & 12.6 & 20.1 & 10.8 & 23.3 & 11.2 & 11.8 & 14.0 \\
GPS & SceneVerse & \best{15.9} & 12.6 & 4.2 & 1.9 & \best{26.5} & \best{14.4} & \best{16.5} & 16.3 \\ %
\midrule
\oursmodel & \ourdataset & 12.1 & 12.4 & 51.5 & 35.7 & 17.5 & 7.3 & 8.5 & 12.6 \\ %
\oursmodel & \ourdatasetlarge & \second{14.2} & \best{12.9} & \best{66.3} & \best{42.1} & 22.9 & \second{12.0} & \second{14.2} & \best{18.2} \\ %
\bottomrule
\end{tabular}
}
\caption{Accuracy and F1 score (\%) on ViGiL3D (ScanNet scenes) and Multi3DRefer using predicted boxes from Mask3D~\citep{schult2023mask3d}.
}
\label{tab:model-evaluation-all-pred-vigil3d-multi3drefer}
\end{table*}

  \begin{table}[tb]
\centering
{
\begin{tabular}{@{}l rr r rrr@{}}
\toprule
& \multicolumn{2}{c}{\oursshortman} & ScanRefer & \multicolumn{3}{c}{Multi3DRefer} \\
\cmidrule(l){2-3} \cmidrule(l){4-4} \cmidrule(l){5-7}
& Acc/GT & F1/GT & Acc@GT & zt/F1 & st/F1 & mt/F1 \\
\midrule
\oursmodel & 14.0 & 15.6 & 44.3 & 42.6 & 18.5 & 22.2 \\
- dense & 14.3 & \best{15.7} & \best{44.4} & \best{51.8} & 16.6 & 20.3 \\
- anchor & \best{15.2} & 15.5 & 43.3 & 30.8 & \best{21.8} & \best{24.9} \\
- dense, anchor & \best{15.2} & 15.3 & 42.0 & 44.4 & 19.4 & 24.1 \\
- gen. obj features & \best{15.2} & 14.9 & 39.5 & 55.2 & 14.9 & 18.5 \\
- pretraining & 13.5 & 13.0 & 24.0 & 53.2 & 7.6 & 12.8 \\

\bottomrule
\end{tabular}
}
\caption{\textbf{Ablation study} of model changes of accuracy and F1 score (\%) on \oursshortman, ScanRefer~\citep{chen2020scanrefer}, and Multi3DRefer~\citep{zhang2023multi3drefer}. Metrics are computed using GT boxes. All models were trained on \ourdataset using only ScanNet scenes, and each row shows the performance when removing a change. We find the greatest improvements come from pretraining, generalized object features, and some form of anchor or dense loss, though neither loss appears significantly better than the other.
}
\label{tab:model-ablation}
\end{table}

  \begin{figure}[ht]
  \centering
  \setkeys{Gin}{width=\linewidth}
  \begin{tabularx}{\textwidth}{p{1in} Y@{\hspace{1mm}} | Y@{\hspace{1mm}} Y@{\hspace{1mm}} Y@{\hspace{1mm}} Y@{\hspace{1mm}} Y}
    & GT & ZSVG3D & 3D-VisTA & 3D-GRAND & GPS & V3DM \\
    A black bag is in front of a white trash can on the floor. & \includegraphics{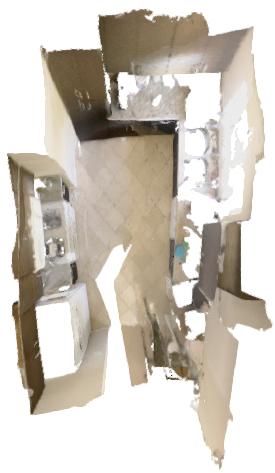} & \includegraphics{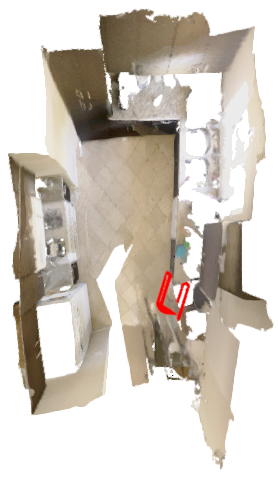} & \includegraphics{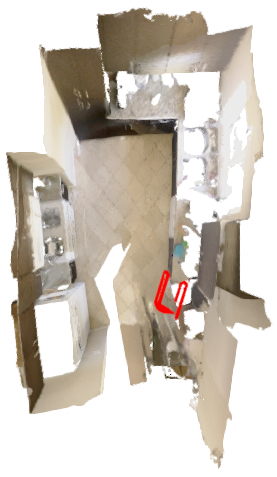} & \includegraphics{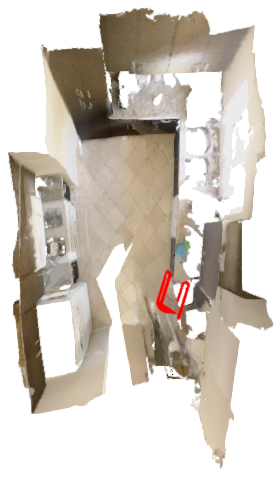} & \includegraphics{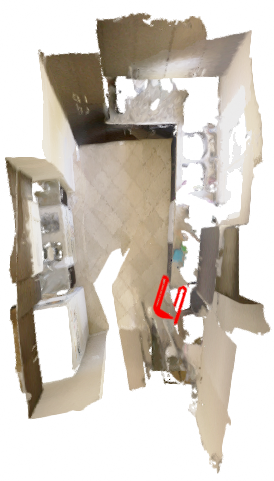} & \includegraphics{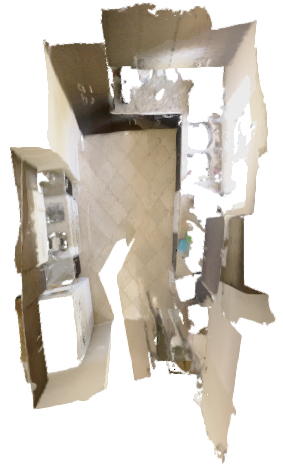} \\
    If facing the wall with the poster with the words, "SIGGRAPH", get all of the tables on your right. & \includegraphics{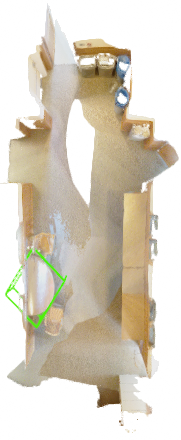} & \includegraphics{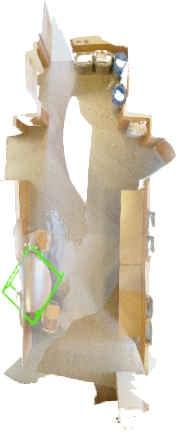} & \includegraphics{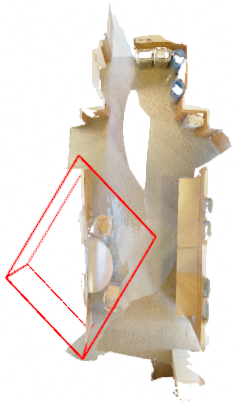} & \includegraphics{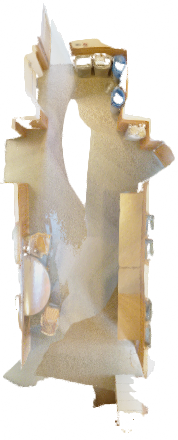} & \includegraphics{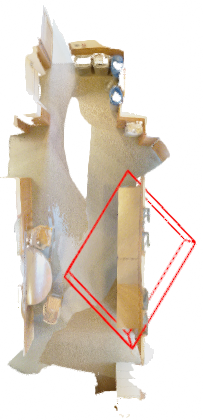} & \includegraphics{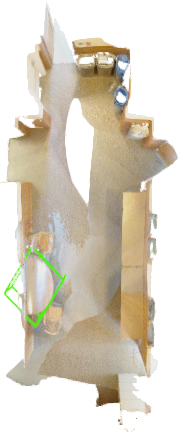} \\
    These are all the long, soft places in the room. & \includegraphics{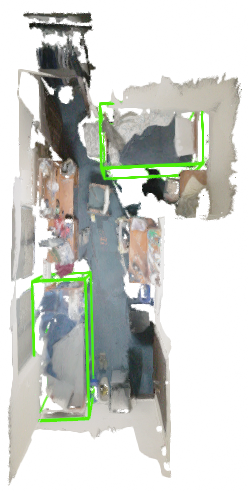} & \includegraphics{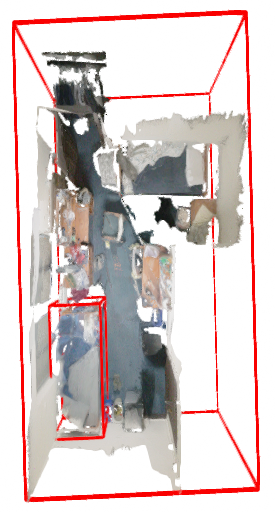} & \includegraphics{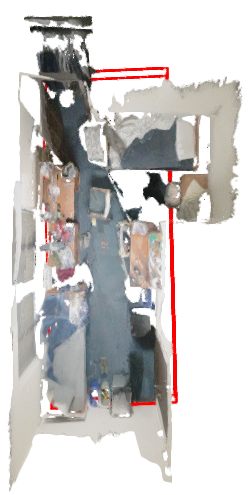} & \includegraphics{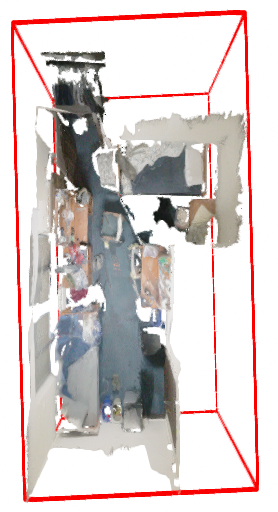} & \includegraphics{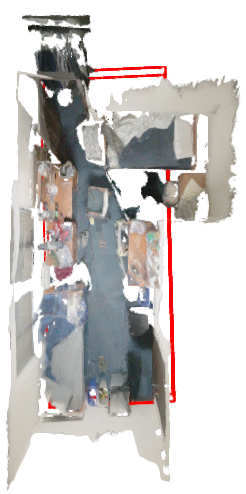} & \includegraphics{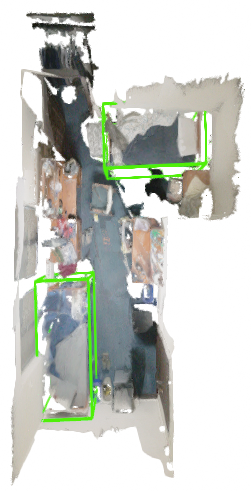} \\
  \end{tabularx}
\captionof{figure}{\textbf{Examples.} \oursmodel trained on \oursshort is able to predict correctly (\textcolor{green}{green}) on challenging zero-, single-, and multi-target queries from \oursshortman over existing methods, which often identify the wrong objects (\textcolor{red}{red}).}
\label{fig:examples}
\end{figure}

  We describe additional details for our model and ablation experiments.

  \subsection{Training Objectives}

    To train our model for 3DVG, we use a combined loss consisting of the target grounding loss \losstarget, the dense alignment loss \lossdense, and the anchor loss \lossanchor.
    The complete loss is given by $\loss = \losstarget + \beta\, (\lossdense + \lossanchor)$,
    where $\beta$ is a fixed hyperparameter.
    We describe each of the losses in detail below.
    
    \mypara{Target Grounding Loss.} We apply an MLP to $f_{1:M}^S$ to get scalar scores $p_{1:M}$ for each object. As we allow for zero, one, or multiple targets, we apply a sigmoid function and binary cross-entropy to independently evaluate each object:
    
    {\small
    \begin{equation}
        \losstarget = -\frac{1}{M}\sum_{i=1}^M \left[t_i \log \sigma(p_i) + (1 - t_i)\log(1 - \sigma(p_i))\right].
    \end{equation}
    }
    
    \mypara{Dense Alignment Loss.} We apply a dense alignment loss between object and text features to align text tokens corresponding to noun phrases to their corresponding referents. We compute this using a focal loss applied to an averaged multihead attention map between the fused object and text features, where a text token embedding can align to multiple objects or no objects, and likewise an object can align to multiple text tokens or none. Letting $A_{ij} = 1$ if object $o_i$ is referred to by text token $t_j$ and 0 otherwise, the loss is formulated as
    {\small
    \begin{equation}
        \ell_{da}(i, j) = 
        - \Big[ 
            A_{ij} \, \log \sigma\!\big(\tau (f_i^S \cdot f_j^T)\big) + \, \big(1 - A_{ij}\big) \, \log \big(1 - \sigma\!\big(\tau (f_i^S \cdot f_j^T)\big)\big)
        \Big]
    \end{equation}
    \begin{equation}
        p_{ij} = \Big[
          A_{ij} \sigma(\tau (f_i^S \cdot f_j^T)) 
          + (1 - A_{ij})(1 - \sigma(\tau(f_i^S \cdot f_j^T)))
        \Big]
    \end{equation}
    \begin{equation}
        \lossdense = \sum_{i=1}^M \sum_{j=1}^T \alpha_{ij} \ell_{da}(i, j) (1 - p_{ij}))^{\gamma}
    \end{equation}
    }
    
    where $\alpha_{ij} = \alpha A_{ij} + (1-\alpha)(1 - A_{ij})$.

    \mypara{Anchor loss.} Following \cite{abdelreheem2024scanents3d}, given the fused object embeddings, during training we learn an MLP to predict the anchor objects referenced in the VG query. Similar to \losstarget, we apply binary cross-entropy as follows:
    {\small
    \begin{equation}
        \lossanchor = -\frac{1}{M}\sum_{i=1}^M \left[a_i \log \sigma(p_i) + (1 - a_i)\log(1 - \sigma(p_i))\right].
    \end{equation}
    }
    Lastly, we weight \lossdense and \lossanchor against \losstarget using $\beta = 3$.

  \subsection{Training Details.}
  \label{appendix:model-training-details}
We train \oursmodel from the pretrained weights of 3D-VisTA~\citep{zhu20233d} for up to 100 epochs or until convergence, using similar training settings to 3D-VisTA~\citep{zhu20233d}, including a learning rate warmup and AdamW optimizer~\citep{loshchilov2017decoupled}, for fair comparison. 
We use a single A40 GPU, running over 3 days.
For training and inference, we leverage ground truth object segmentations for object proposals to isolate the comparison of the methods themselves, though a 3D object segmentation model can also be used to pre-segment the scene.

  \subsection{Model Experiments}
  \label{appendix:model-experiments}

  For each of the three methods tested, we used pretrained checkpoints, following \cite{wang2024vigil3d}. 
  For \emph{ZSVG3D}, we utilized GPT-4o as the LLM and CLIP (ViT-B/16)~\citep{radford2021learning} for the LOC module. 
  \emph{3D-VisTA} was pretrained on ScanScribe and fine-tuned on ScanRefer for visual grounding. We extended the point cloud sequence length to 120 to support larger scenes. 
  \emph{3D-GRAND} was pretrained on the 3D-GRAND dataset and uses Llama-2-7b as its LLM. We used the \texttt{merged\_weights\_grounded\_obj\_ref} checkpoint provided by the authors.

  \mypara{Mask3D predictions.} We show the results when running on Mask3D predictions instead of GT in \cref{tab:model-evaluation-all-pred-scanrefer} for ScanRefer and \cref{tab:model-evaluation-all-pred-vigil3d-multi3drefer} for ViGiL3D (ScanNet scenes only) and Multi3DRefer. We find that the performance of \oursmodel is still competitive with other models and performs better than its most comparable baseline in 3D-VisTA.

  \mypara{Ablation study.} We compare performance when incorporating generalized object features (\ie no GloVe vectors) and when using the anchor and dense alignment loss functions. We also compare the performance of pretraining on ScanScribe against training from scratch. We train on the \ourdataset descriptions of ScanNet scenes. The results are shown in \cref{tab:model-ablation}. We find that the combination of model changes and loss functions largely improves performance on all benchmarks, with major gains coming from pretraining and open-vocabulary and modest gains from some type of anchor-based loss. We hypothesize that the dense loss is not as effective given the complexities of modeling exactly which objects are linked to each referent, but further iteration of the loss function and dense annotations may allow the model to better leverage and predict dense object alignment in future work.

\section{Scientific Artifacts}
  \label{appendix:attribution}

  The licenses used in this paper include the following: 
  ScanNet (terms of use\footnote{\url{http://kaldir.vc.in.tum.de/scannet/ScanNet_TOS.pdf}}),
  3RScan (MIT), 
  MultiScan (MIT),
  3D-GRAND (CC BY 4.0), 
  3D-VisTA and ScanScribe (MIT), 
  SceneVerse (terms of use \footnote{\url{https://drive.google.com/file/d/14Ji7PLOKsAxrXpxV6EWLsQGjzcEuk35N/view}}), 
  ViGiL3D (MIT), 
  ZSVG3D (N/A), 
  OpenAI (terms of use\footnote{\url{https://openai.com/policies/terms-of-use/}}), 
  Qwen VL (terms of use\footnote{\url{https://github.com/QwenLM/Qwen-VL/blob/master/LICENSE}}), 
  InternVL (MIT),
  SoM prompting (MIT),
  and \texttt{diversity} (Apache 2.0). 
  We follow the intended use of all the licenses in the paper and reported our intended usage in the terms as appropriate.

  LLMs, including ChatGPT~\citep{achiam2023gpt}, Qwen-3 VL~\citep{qwen3technicalreport}, and InternVL~\citep{wang2025internvl3_5}, were used in \oursshort and in some of the baseline methods. We also used them as assistive tools for generating code and researching methodologies.

\end{document}